%% file: acl_latex.tex
\pdfoutput=1

\documentclass[11pt]{article}

\usepackage{acl}

\usepackage{times}
\usepackage{latexsym}

\usepackage[T1]{fontenc}

\usepackage[utf8]{inputenc}

\usepackage{microtype}

\usepackage{inconsolata}

\usepackage{csquotes}
\usepackage{algorithm}
\usepackage{algpseudocode}
\usepackage{multirow}
\usepackage{booktabs}
\usepackage{microtype}
\usepackage{colortbl}
\usepackage{subcaption}
\usepackage{twemojis}
\usepackage{tikz}
\usepackage{threeparttable}
\usepackage{ragged2e}
\usepackage{soul}
\usepackage{enumitem}
\usepackage{etoolbox}
\AtBeginEnvironment{tcolorbox}{%
\setlist[itemize]{nosep,
                 leftmargin=*,
                 label=\textbullet,
                 before=\begin{minipage}[t]{\linewidth}, 
                 after=\end{minipage}\medskip}                   
                            }

\usepackage[most]{tcolorbox}
\tcbuselibrary{raster}

\newcommand{\bug}{\scalebox{1.75}{\twemoji{beetle}}}
\newcommand{\bugone}{\scalebox{1.25}{\twemoji{beetle}}\textsubscript{1}}
\newcommand{\bugtwo}{\scalebox{1.25}{\twemoji{beetle}}\textsubscript{2}}
\newcommand{\bugthree}{\scalebox{1.25}{\twemoji{beetle}}\textsubscript{3}}
\newcommand{\bugall}{\scalebox{1.25}{\twemoji{beetle}}\textsubscript{1,2,3}}
\newcommand{\smallbug}{\scalebox{1.25}{\twemoji{beetle}}}
\newcommand{\correct}{\scalebox{1.75}{\twemoji{check mark button}}}
\newcommand{\smallcorrect}{\scalebox{1.25}{\twemoji{check mark button}}}

\newcommand{\smallwrong}{\scalebox{1.25}{\twemoji{cross mark}}}

\newcommand{\idea}{\scalebox{1.25}{\twemoji{light bulb}}}
\newcommand{\code}{\scalebox{1.25}{\twemoji{laptop}}}
\newcommand{\results}{\scalebox{1.25}{\twemoji{bar chart}}}
\newcommand{\school}{\scalebox{1.25}{\twemoji{school}}}
\newcommand{\torchimpl}{TorchAudio}

\newcommand{\mg}{\textcolor{black}}

\title{When Good and Reproducible Results are a Giant with Feet of Clay:\\ The Importance of Software Quality in NLP}

\author{
Sara Papi\textsuperscript{\idea\results\school}, Marco Gaido\textsuperscript{\idea\results}, Andrea Pilzer\textsuperscript{\code}, Matteo Negri\textsuperscript{\results} \\
  \results Fondazione Bruno Kessler \\
  \school University of Trento \\
  \code NVIDIA \\
  \texttt{\{spapi,mgaido,negri\}@fbk.eu,apilzer@nvidia.com}}

\begin{document}
\maketitle
\begin{abstract}
Despite its crucial role in research experiments, code correctness is often presumed solely based on the perceived quality of results. This assumption, however, comes with the risk of erroneous outcomes and, in turn, potentially misleading findings. To mitigate this risk, we posit that the current focus on reproducibility should go hand in hand with the emphasis on software quality. We support our arguments 
\mg{with} a case study in which we identify and fix three bugs in widely used implementations of the state-of-the-art Conformer architecture.
\mg{Through experiments on speech recognition and translation in various languages, we demonstrate that the presence of bugs does not prevent the achievement of good and reproducible results, which however can lead to incorrect conclusions that potentially misguide future research.} 
As countermeasures, we release \texttt{pangoliNN}, a library dedicated to testing neural models, and propose a Code-quality Checklist, with the goal of promoting coding best practices and improving software quality within the NLP
community.
\end{abstract}

\section{Introduction}
\label{sec:intro}
In the field of natural language processing (NLP), as well as in broader contexts, the validity and soundness of research findings are typically upheld by \enquote{\textit{establishing consistency of results versus existing implementations, standard benchmarks, or sanity checks via statistically significant experimental results}} \citep{Rozier-2014-repro}.
Embracing these recommendations as the exclusive criteria for validating scientific credibility,
the 
research community
has recently devoted significant attention to 
the reproducibility 
\citep{dodge-etal-2019-show,branco-etal-2020-shared,belz-etal-2021-reprogen,10.1162/coli_a_00448,the_turing_way_community_2022} and  the soundness of experimental settings and comparisons \citep{denkowski-neubig-2017-stronger,dror-etal-2018-hitchhikers,marie-etal-2021-scientific}.
Specifically, in response to evidence indicating the absence of these aspects in many research papers \citep{NEURIPS2019_c429429b} and to mitigate the so-called  
``\textit{reproducibility crisis}'' \citep{Baker2016},\footnote{The term ``\textit{reproducibility crisis}'' refers to the increasing difficulties reported by scientists in replicating others' and own works \citep{prinz2011believe,Gundersen_Kjensmo_2018,10.1162/coli_a_00330,Chen2019,Gundersen_2019}, also in the specific context of NLP \citep{wieling-etal-2018-squib,marie-etal-2021-scientific,narang-etal-2021-transformer,gehrmann2022repairing,arvan-etal-2022-reproducibility-code,belz-etal-2021-systematic,belz-etal-2022-quantified,belz-etal-2023-non}.} top-tier conferences have introduced dedicated checklists and targeted questions in the reviewing forms \citep{pineau2021improving,rogers-etal-2021-just-think}.

However, 
a fundamental question remains regarding the initial assumption: \textbf{are reproducibility and thorough evaluation against robust baselines
sufficient to ensure the soundness of a research finding?}
%
According to \citet{Peng-2011-reproducible}, reproducibility alone
does 
not
\enquote{\textit{guarantee the quality, correctness, or validity of the published results}} since the code employed to produce them may not accurately execute its intended purpose.
This entails inherent risks, as flawed code that produces good and easily reproducible results can propagate as the foundation for further research, ultimately leading to further unreliable and potentially misleading findings \citep{McCullough-2008-replicable}.


Expanding on these observations, this paper is a call to action, underpinned by empirical evidence, 
to bolster the dependability of NLP findings by complementing
current initiatives toward 
reproducibility and experimental soundness
with equal emphasis on \textit{software quality}. 
To this 
end, we adopt as a reference framework the principles of software quality assurance (SQA -- \citealt{Buckley-1984-sqa,tripathy2011software}), which have so far been overlooked by our community.
Building on this foundation, we contribute as follows:

\begin{enumerate}[leftmargin=14pt]

\item We examine the extent to which research works consider the attributes studied in the SQA field
(\S\ref{sec:sqa}), 
highlighting
that code correctness has been neglected by the NLP community thus far (\S\ref{sec:core-idea});
    
\item Through a case study on open-source implementations of the widespread Conformer architecture \citep{gulati20_interspeech}, we
show that:
\begin{itemize}[leftmargin=8pt]
\item[-] At least one impactful bug is present in all the analyzed implementations (\S\ref{subsec:analysis});
\item[-] 
Such bugs do not
prevent from achieving
good and
reproducible results that outperform
other architectures
in speech recognition and translation across different
language 
pairs
(\S\ref{subsec:impact_bug});
\item[-] 
Undetected bugs can lead to erroneous conclusions when evaluating new techniques
(\S\ref{subsec:impact_code}).
\end{itemize}
\item We release a bug-free implementation of
Conformer,\footnote{\label{foot:sw_release}Availabe at \url{https://github.com/hlt-mt/FBK-fairseq/} under the Apache 2.0 License.} 
along with all the pre-trained models;
\item  \mg{We promote code correctness and software quality by releasing \texttt{pangoliNN},\footnote{Availabe at \url{https://github.com/hlt-mt/pangolinn/} under the Apache 2.0 License.}
a library featuring easily-usable unit tests to enforce the proper behavior of neural models (\S\ref{subsec:pangolinn}), and proposing the integration into current conferences checklists of a Code-quality section, which would focus on coding best practices (\S\ref{subsec:checklist}).}
\end{enumerate}

\section{SQA and Research}
\label{sec:sqa}
\input{sections/sqa}

\section{Research Code Quality Evaluation}
\label{sec:core-idea}
\input{sections/code_quality_eval}

\section{The Case Study}
\label{sec:case-study}

\input{sections/case_study}

\section{Increasing Research Code Correctness}
\label{sec:checklist}
\input{sections/proposal}

\section{Conclusions}

In parallel with the current efforts to enhance the reproducibility of NLP research, this paper urges similar actions targeting the improvement of research software quality, underscoring its importance for the reliability of research findings.
%
In comparison to the attention given to  
reproducibility within our community,
we observed the predominant neglect of code correctness and elaborated on the risks associated with assessing soundness solely on the basis of experimental results.
As we empirically demonstrated through a case study involving the widespread Conformer architecture, such risks include the potential of drawing misleading conclusions 
from positive 
results obtained 
using
flawed code.
%
%
As a countermeasure, 
besides releasing a corrected Conformer implementation, we created the \texttt{pangoliNN} Python package to facilitate testing neural models and proposed the adoption of a \enquote{Code-quality Checklist} 
aimed at fostering coding best practices.
While we acknowledge that these solutions are not a panacea, their purpose is to raise awareness within the NLP community about the importance of software quality. We hope that our 
endeavor will inspire a collective commitment 
to developing
%
high-quality and reliable code.

\section*{Acknowledgements}
We acknowledge the support  of the PNRR project FAIR -  Future AI Research (PE00000013),  under the NRRP MUR program funded by the NextGenerationEU.
We acknowledge the CINECA award under the ISCRA initiative, for the availability of high-performance computing resources and support.

\section*{Limitations}
To back up our call to action toward the adoption of coding best practices aimed at fostering correctness and improving the quality of the developed software, we presented a case study involving the use of the Conformer architecture in the two most popular speech processing tasks: speech recognition and translation. Although the effects of the presence of bugs might be found also in other  scenarios, such as  text-to-speech, speech emotion recognition, spoken language understanding, and speech separation, we did not cover them in this paper.  While the undesired effect of the bugs we isolated (and corrected) was empirically demonstrated, extending the analysis to other research areas would be a natural extension of our study, which could provide a more comprehensive understanding of the impact of the identified bugs on the broader NLP community working on speech-related tasks.

Moreover, in our case study, we examined the open-source implementations of Conformer, in which we identified three types of bugs related to the Convolution Module, Initial Subsampling, and Positional Encodings. While we found efficient solutions for the first two bugs, for the last one our fix introduces a significant overhead. As a result, the implementation we release, although correct, increases the training time of the models. We are confident that, by open-sourcing our code, the community will soon find a way to optimize it and overcome this limitation, capitalizing on our findings and spreading the use of more reliable versions of a state-of-the-art architecture.

\bibliography{custom}

\appendix

\begin{figure*}[!t]
    \centering
    \includegraphics[width=0.95\textwidth]{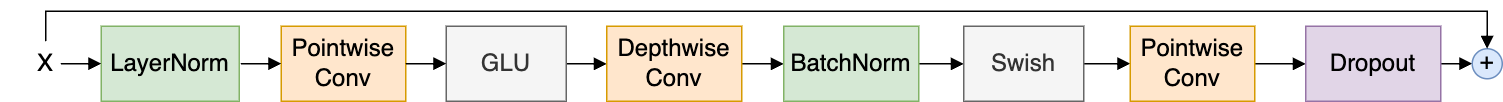}
    \caption{Convolution module in the Conformer encoder layer. Convolutional blocks are 1D convolutions.}
    \label{fig:conf_conv}
\end{figure*}

\section{Conformer in ASR and ST}
\label{subsec:speech_background}

ASR is the task in which an audio containing speech content is transcribed in its original language. In ST, instead, the source audio is translated into text in a different language. Nowadays, both tasks are commonly performed with end-to-end (or direct) models \citep{pmlr-v32-graves14,Chorowski-2014-asr,berard_2016,weiss2017sequence}, whose architecture is based on the Transformer~\citep{NIPS2017_3f5ee243}. The Transformer has been adapted to work with audio inputs~\citep{8462506,gangi19_interspeech} by introducing two convolutional layers that shrink the length of the input sequence by a factor of $4$, so as to reduce the otherwise excessive memory requirements. More recently, \citet{gulati20_interspeech} proposed the Conformer: a novel architecture with a modified encoder that led to significant improvements in both ASR and ST \citep{inaguma2021non}.

The changes introduced in the Conformer encoder layer structure can be summarized as follows:
\textit{i)} relative sinusoidal positional encodings \citep{dai-etal-2019-transformer} are introduced in the self-attention for improved generalization with respect to varying input lengths;
\textit{ii)} the FFN sublayer is replaced by two FFNs that wrap the self-attention, inspired by the Macaron-Net~\citep{lu-et-al-2016-macaron-net};
\textit{iii)}~a convolution module 
(Figure \ref{fig:conf_conv}) is added
after the self-attention, before the second FFN layer.
The convolution module, which is wrapped in a residual connection, applies layer normalization, followed by
a pointwise convolution that doubles the dimension of the feature vector, which
is restored to its original size by a Gated Linear Unit (GLU) activation function~\citep{Dauphin-2017-glu}. Then, a depthwise convolution with 31 kernel size is applied before a batch normalization \citep{ioffe-2015-batchnorm}, followed by the Swish activation function \citep{swish-2017}, and another pointwise convolution. Lastly, a dropout module \citep{Srivastava-2014-dropout} randomly
zeroes out
a percentage of the
features
to prevent the network from overfitting.

\section{Experimental Settings}
\label{app:expsett}

Our Conformer-based architecture is composed of 12 Conformer \citep{gulati20_interspeech} encoder layers and 6 Transformer \citep{NIPS2017_3f5ee243} decoder layers, with 8 attention heads each. Embedding size is set to 512 and hidden neurons in the feed-forward layers to 2,048, with a total of 114,894,730 
model parameters. Dropout is set to 0.1 for feed-forward, attention, and convolution layers. The kernel size of the Convolution Module is set to 31 for both point-wise and depthwise convolutions.  We train all the models using Adam \citep{journals/corr/KingmaB14} optimizer (betas $(0.9, 0.98)$) and label-smoothed cross-entropy (LSCE) loss (smoothing factor 0.1). We also use an auxiliary Connectionist Temporal Classification or CTC loss \citep{Graves2006ConnectionistTC} during training to ease convergence and obtain competitive results without pre-training the encoder with that of an ASR model \citep{gaido-etal-2022-efficient}. The auxiliary loss is summed to the LSCE with 0.5 relative weight.  The learning rate is set to $2\cdot10^{-3}$ with Noam scheduler \citep{NIPS2017_3f5ee243} and 25k warm-up steps.  The vocabularies are based on SentencePiece models 
\citep{kudo-richardson-2018-sentencepiece}
with size 5,000 \citep{inaguma-etal-2020-espnet} for the English source 
and 8,000 \citep{di-gangi-etal-2020-target} for the ST target languages.
We set 100k maximum updates with early stopping after 10 epochs without loss decrease on the dev set and average 5 checkpoints around the best (best, two preceding, and two following). All trainings are performed with 40k tokens as batch size and 4 as update frequency on two GPUs. All other settings are the default of Fairseq-ST~\citep{wang2020fairseqs2t}, which we forked as a base of our implementation. SpecAugment \citep{Park2019} is applied during training, while utterance-level Cepstral mean and variance normalization is performed both at training and inference time. Trainings lasted 18-33 hours depending on the model configuration 
(e.g., with or without the fixes)
and the language pair due to the different sizes of the training data.

\section{CTC compression}
\label{app:ctc_compr}

CTC compression 
has been proposed to reduce the difference in terms of sequence length between corresponding audio and text representations.
In contrast with
fixed reduction methods like max pooling or strided convolutions that apply a predetermined reduction to each sequence, CTC compression leverages
the probability distribution over the source vocabulary augmented with a \texttt{<blank>} symbol produced by the CTC module.
These probabilities are
used to assign a label (the most likely one) to each vector of the sequence and collapse contiguous vectors corresponding to the same label by averaging them.
By dynamically determining which vectors of the audio sequence should be merged, it tries to avoid the mismatch in terms of sequence length with the sub-word 
sequence
of the corresponding 
transcript.

\end{document}

%% file: sections/sqa.tex
Software Quality Assurance (SQA) attributes have been studied for many years \citep{McCall1977FactorsIS,10.5555/48813,10.5555/121025}. Delineated in the ISO 9126 standard \citep{ISO9126}, they were later extended and superseded by ISO 25010 \citep{ISO/IEC2010}, having production code as the main target. However, as they are desirable for any codebase, here we analyze how each attribute has an effect on research code and work.

\textbf{Portability} and \textbf{usability} refer, respectively, to
the possibility of executing the same experiments in diverse hardware or software environments, and the effort required to use the software (i.e., how easy it is to run the code).
As such, they pertain to the reproducibility of a paper, which, according to ACM,\footnote{\url{https://www.acm.org/publications/policies/artifact-review-and-badging-current}} holds when \enquote{\textit{an independent group can obtain the same result using the author's own artifacts}}, a definition that is aligned with those given in NLP \citep{ulmer2022experimental} and other fields \citep{doi:10.1128/mBio.00525-18}.
Regarding these aspects, ample literature already discussed the need to go beyond code openness in research \citep{Chen2019,Trisovic2022}, highlighting the role of proper documentation and validation in
different environments.
However, as many research groups lack access to a wide range of hardware options, we argue that research works can hardly target portability due to the significant economic and human resources it requires.
On the contrary, proper documentation of the code is a reasonable demand and, in addition to increasing reproducibility, it facilitates code reuse and
adoption for other works.

Code reusability also pertains to the
\textbf{maintainability} attribute, which denotes the effort required for implementing targeted modifications. 
Alongside comprehensive documentation, software maintainability hinges on code structure,
i.e. the organization of the software into building blocks 
\citep{10.1145/141874.141884,intro_soft_architectures}.
While there is currently no incentive to develop reusable code \citep{barba-2019-praxis},  the research community would greatly benefit in the long term from a commitment to this objective, which would reduce the time spent in replicating prior work and accelerate the implementation of new techniques upon existing code.

The 
expeditiousness 
of testing new methods also depends on the \textbf{efficiency} and \textbf{reliability} of the codebase.
Efficiency refers to the amount of resources a software uses, e.g. the number of GPU hours or VRAM GBs needed for training.
Increasing efficiency constitutes a research direction on its own and can hardly be considered a prerequisite for orthogonal investigations.
Reliability, instead, is the capability of the software to seamlessly operate in all conditions and for a long time: software causing frequent crashes (i.e. terminations due to errors) or whose efficiency is not constant over time is not reliable.
Although both properties would contribute to reducing the environmental footprint of NLP research by avoiding computing-resource wastes or unexpected failures \citep{strubell-etal-2019-energy,Shterionov2023},
a commitment in this direction
is arguably 
an excessive demand for research works not expressly dedicated to it.

Last but not least, \textbf{functionality} or \textbf{functional correctness} (hereinafter: correctness)  pertains to the \enquote{\textit{extent to which a program satisfies its specifications}}  \citep{McCall1977FactorsIS}. In research, this holds when the code exactly performs the operations described in a paper, thereby establishing the validity of the reported findings.
Achieving correctness requires the creation and execution of tests, as they are the sole mechanism that guarantees the correct behavior of software.
For example, when designing a causal model (i.e. a model that cannot look at future elements in 
the input sequence),
researchers should test that the model predictions always remain unaffected by future elements. If a bug breaks the causality property, any observed gains may not stem from the proposed solutions but from undue access to forbidden information.
It is worth emphasizing that the validity of these tests expires after any code alteration, regardless of its apparent relevance. Therefore, tests should be executed after each modification to  ensure correctness and, in turn, 
the trustworthiness of the findings.

In summary, we have observed that, in the context of research software, \textit{i)} portability and usability support reproducibility, \textit{ii)} maintainability promotes reusability, \textit{iii)} efficiency and reliability reduce environmental costs, and \textit{iv)} correctness plays a 
crucial
role in ensuring trustworthy findings and research soundness. 
However, despite its importance, we show in the next section that the research community has largely neglected correctness, 
focusing primarily on reproducibility.

%% file: sections/code_quality_eval.tex
To assess the level of consideration given to the above SQA attributes within the NLP research community, we examined their inclusion in the review forms of top-tier conferences and journals in the field, namely: 
*ACL (i.e., AACL-IJCNLP, ACL, EACL, EMNLP, NAACL),\footnote{Since EACL 2024, *ACL conferences adopt ARR only.} ACL Rolling Review (ARR), ICASSP, ICML, ICLR, Interspeech, NeurIPS, and TACL.
We specifically focused on reproducibility (as a proxy of portability and usability) and correctness. Table \ref{tab:nlp} shows the results.

Most of the venues (5 out of 8) include an explicit score for reproducibility and NeurIPS mentions it among the factors contributing to the overall recommendation score. Reproducibility is commonly evaluated through dedicated checklists\footnote{E.g., NeurIPS (\url{https://neurips.cc/Conferences/2021/PaperInformation/PaperChecklist}), AAAI (\url{https://aaai.org/Conferences/AAAI-22/reproducibility-checklist}), and ARR (\url{https://aclrollingreview.org/responsibleNLPresearch}).} that mainly focus on the detailed descriptions of the hyperparameters and the software/hardware environment (while disregarding whether different hardware/software is supported, i.e. portability, which seems reasonable as seen in \S\ref{sec:sqa}).
Accordingly, these checklists are not strictly related to SQA, although they do include recommendations for proper code documentation, which is related to the software usability and maintainability.

Correctness is instead mentioned in fewer forms (3 out of 8). When present, its definition varies and is not explicitly related to the code: at ICLR and ICASSP, the scope of the term is not clearly defined, while in TACL it is included in the broader concept of \textit{soundness} of the experiments/results.
In this result-oriented definition, soundness pertains to assessing the significance of results (\textit{are the reported improvements robust to statistical fluctuations?}) with respect to either the state of the art (\textit{are the results competitive with those reported in recent literature?}) or strong baselines.
Soundness is also assessed at NeurIPS and ICML but, again, code correctness is never explicitly mentioned.
Notably, the Interspeech form contains a \enquote{Technical Correctness} score, which however refers to the reproducibility of the paper (\enquote{\textit{are enough details provided to be able to reproduce the experiments?}}).
In general, when considered, software is explicitly evaluated only in terms of accessibility (\textit{is the code released open-source?}) and potential usefulness (\textit{will the research community benefit from the use of the software?}).
For instance, the \enquote{Software} score in the ARR form only refers to the usefulness and documentation of newly-released code rather than to its correctness, thus being again more related to its usability and maintainability.

\begin{table}[!t]
    \centering
    \small
    \setlength{\tabcolsep}{8pt}
    \begin{tabular}{l||c|c}
    \specialrule{.1em}{.05em}{.05em} 
        \textbf{Venue} & \textbf{Reproducibility} & \textbf{Correctness} \\
        \specialrule{.1em}{.05em}{.05em} 
        *ACL &  \smallcorrect{} & \smallwrong{} \\
        ARR & \smallcorrect{} & \smallwrong{} \\
        ICASSP & \smallwrong{} & \smallcorrect{} \\
        ICML & \smallwrong{} & \smallwrong{} \\
        ICLR & \smallcorrect{} & \smallcorrect{} \\
        Interspeech & \smallcorrect{} & \smallwrong{} \\
        NeurIPS & \smallwrong{} & \smallwrong{} \\
        TACL & \smallcorrect{} & \smallcorrect{} \\
    \specialrule{.1em}{.05em}{.05em} 
    \end{tabular}
    \caption{Reproducibility and correctness in the review forms of major NLP conferences/journals.}
    \label{tab:nlp}
\end{table}

We can conclude that, unlike reproducibility-related SQA attributes, correctness is largely neglected in favor of a result-based evaluation of soundness.
From the researchers' perspective, this entails the risk of basing future work on unreliable software that yields high and easily reproducible results but lacks guarantees of its correctness.
This risk, in turn, can lead to misleading findings \citep{McCullough-2008-replicable}. In the next section, we present a concrete instance of this problem with a case study analyzing open-source implementations of the widespread Conformer architecture.

%% file: sections/case_study.tex
In our case study, we examine the Conformer \citep{gulati20_interspeech} architecture --
the state-of-the-art solution for speech processing tasks \citep{conformer-sota,ma2021end,conformer-sota2,conformer-sota3} such as automatic speech recognition (ASR) and speech-to-text translation (ST) -- whose rapid and wide adoption is
evidenced by 
more than 2,000
citations since 2020.\footnote{Source: Google Scholar --  November 15th, 2023.}

\mg{In the following, we first analyze the Conformer implementation of six widely-used open-source codebases, showing that they all contain at least one bug (\S\ref{subsec:analysis}). Then,}
%
%
%
through extensive experiments on the two tasks and on eight language pairs, we demonstrate that the presence of bugs can be hidden by good -- but incorrect -- results (\S\ref{subsec:impact_bug}),
consequently leading to erroneous conclusions (\S\ref{subsec:impact_code}).
\mg{An introduction of the ASR and ST tasks object of our study, along with an overview of the Conformer architecture is provided in Appendix \ref{subsec:speech_background}.}

\subsection{Analysis of the Codebases}
\label{subsec:analysis}

We analyze the behavior of the open-source implementations of the Conformer by systematically varying a parameter that should not affect the results: the inference batch size (IBS).
With high IBSs, many samples are collected in the same batch, allowing for their parallel processing on GPU to reduce the overall computational cost.
When samples of different lengths are collected in the same batch -- a frequent situation in speech tasks, where the input length largely varies -- the input sequences are brought to the same dimension by filling them with padding. 
Since with correct implementations the results are independent of the presence of padding (and, therefore, of the IBS), research papers usually include only the training batch size (which, instead, is an important hyperparameter for the stability of the training). However, as we demonstrate in this section, the bugs present in the Conformer implementations undermine the above assumption.

We studied six widely-used repositories, namely: Fairseq-ST~\citep{wang2020fairseqs2t}, ESPnet-ST~\citep{inaguma-etal-2020-espnet}, NeMo~\citep{kuchaiev2019nemo}, SpeechBrain~\citep{speechbrain}, an open source codebase named \enquote{Conformer},\footnote{\url{https://github.com/sooftware/conformer}} and \torchimpl{}~\citep{yang2021torchaudio}.
We discovered that all these implementations return different results with different IBSs, showing that the presence of padding incorrectly alters the results.\footnote{We emphasize that our intention is not to single out the shortcomings of individual libraries. Conversely, we are extremely thankful for the invaluable contribution they represent to our community. Our analysis is only intended to further improve the reliability of codes and, consequently, of the experimental results, which we believe is of utmost importance.}
Upon inspection of the codes, we isolated three bugs associated with padding handling in:
Conformer Convolutions (\bugone), Initial Subsampling (\bugtwo), and Positional Encodings (\bugthree).



\begin{figure}[!tb]
\centering
\renewcommand*\thesubfigure{\arabic{subfigure}} 
     \begin{subfigure}[b]{0.45\textwidth}
        \centering
         \includegraphics[width=0.35\textwidth]{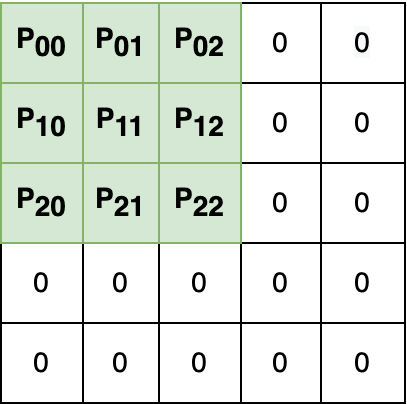}
         \caption{Before shifting, the Relative PE matrix ($P_{00},...,P_{22}$) is padded (zero values).}
     \end{subfigure}
     \par\medskip
     \begin{subfigure}[b]{0.46\textwidth}
        \centering
         \includegraphics[width=0.35\textwidth]{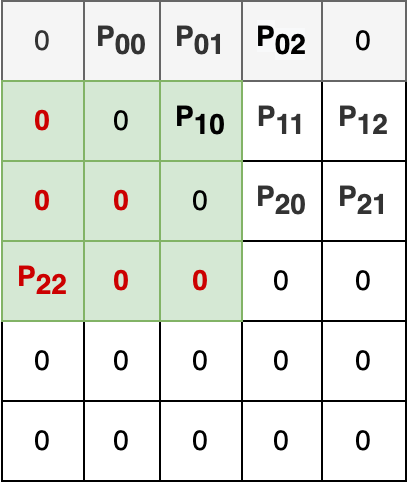}
         \caption{When relative shift is applied to the Relative PE matrix without considering padding, some values of the padding area (in \textcolor{red}{\textbf{red}}) are incorrectly 
         moved to
         the non-padding area.}
     \end{subfigure}
     \par\medskip
     \begin{subfigure}[b]{0.46\textwidth}
        \centering
         \includegraphics[width=0.35\textwidth]{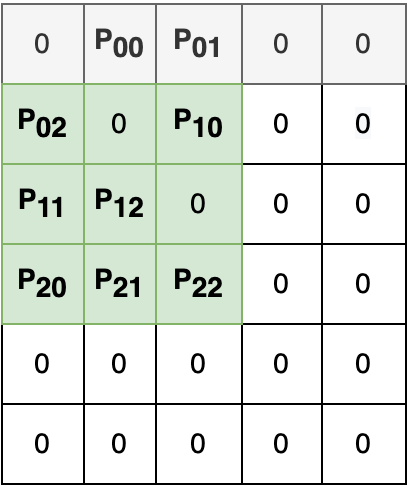}
         \caption{When relative shift is applied to the Relative PE matrix considering padding, the values $P_{00},...,P_{22}$ are not moved to the padding area.}
     \end{subfigure}
    \caption{Example of relative shift operation starting from a Relative PE matrix containing padding (1), both considering a codebase with \bugthree{} (2) and without (3) bug. The first row is always discarded.}
    \label{fig:relativePEs}
\end{figure}

\paragraph{Conformer Convolutions (\bugone)} The depthwise and pointwise convolutions of the Conformer convolution module do not consider the presence of padding and produce a non-padded output with non-zero values adjacent to the input sample. These values modify the behavior of the subsequent batch normalization and of the other convolutions, leading to incorrect alterations of
the
valid values.

\paragraph{Initial Subsampling (\bugtwo)} The two initial convolutions that subsample the input sequence by a factor of 4 do not consider padding. 
\mg{Hence,}
the second convolution is fed with non-zero values adjacent to the input 
sequence, which are wrongly considered in the 
computation of the last valid elements.

\paragraph{Positional Encodings (\bugthree)} The relative sinusoidal positional encodings (PEs), which are added to the attention matrix, are computed by shifting a sinusoidal matrix. This shifting operation first prepends a zero column to the sinusoidal matrix and then reshapes it so that the last element of the first row becomes the first element of the second row, the last two elements of the second row become the first ones of the third row, and so on. By doing this, this operation assumes that all elements are valid. However, when a sequence is padded, only 
part of the attention matrix is valid (in green in Figure~\ref{fig:relativePEs}.1) and spurious values are moved to the beginning of the next row (Figure~\ref{fig:relativePEs}.2).  In Figure~\ref{fig:relativePEs}, for the sake of clarity of the example, we set to 0 the PE in the padding area. While this is not what happens in practice (as the padding area contains other sinusoidal PEs), it shows that the correct values are discarded and the final matrix significantly differs from the one obtained without padding, which is instead shown in Figure~\ref{fig:relativePEs}.3.

\begin{table}[!t]
\small
    \centering
    \setlength{\tabcolsep}{5pt}
    \begin{tabular}{l|c|c|c}
    \specialrule{.1em}{.05em}{.05em} 
         \textbf{Repository} & \textbf{Conv. Mod.} & \textbf{SubSampl.} & \textbf{Pos. Enc.} \\
         \specialrule{.1em}{.05em}{.05em} 
         Fairseq-ST & \bugone & \bugtwo & \bugthree \\
         ESPnet-ST & \bugone & \bugtwo & \bugthree \\
         NeMo & - & \bugtwo & \bugthree \\
         SpeechBrain & \bugone & \bugtwo & \bugthree \\
         Conformer & \bugone & \bugtwo & \bugthree \\
         \torchimpl{} & \bugone & NA & NA \\
    \specialrule{.1em}{.05em}{.05em} 
    \end{tabular}
    \caption{Bugs present in the analyzed repositories. NA stands for \enquote{Not Applicable}.}
    \label{tab:bug}
\end{table}

In Table \ref{tab:bug}, we report the presence (or absence) of these bugs for each analyzed codebase in its current version.
All the implementations but one (NeMo) are affected by \bugone. Also, all are affected by \bugtwo{} and \bugthree, except for \torchimpl, whose implementation neither includes relative positional encodings in the attention nor the initial sub-sampling convolutional layers.
Having ascertained that all the analyzed implementations contain at least one bug, the next sections will concentrate on their impact on ASR and ST results and, in turn, the related findings.


\subsection{Experimental Settings}
We train and evaluate ASR and ST models on MuST-C
v1.0 \cite{CATTONI2021101155}, which contains parallel speech-to-text data with English (en) as source language and 8 target text languages, namely Dutch (nl), French (fr), German (de), Italian (it), Portuguese (pt), Romanian (ro), Russian (ru), and Spanish (es). 
\mg{For ASR, we use the en-es section (the largest of the corpus).}
For ST, 8 different models are trained, one for each language direction. 
Evaluation is performed on the
tst-COMMON, by computing word error rate (WER) for ASR and BLEU with SacreBLEU \citep{post-2018-call}\footnote{BLEU|\#:1|c:mixed|e:no|tok:13a|s:exp|v:2.0.0} for ST\mg{. We assess statistical significance using bootstrap resampling \citep{koehn-2004-statistical} with 95\% confidence interval.}
\mg{Detailed experimental settings are reported in Appendix \ref{app:expsett}.}

Trainings and inferences were performed on, respectively, two and one A40 GPU(s).
On Ampere GPUs, PyTorch  computes convolutions and matrix multiplications with TensorFloat-32\footnote{\url{https://pytorch.org/docs/stable/notes/cuda.html\#tensorfloat-32-tf32-on-ampere-devices}.} (\textbf{TF32}) tensor cores by default. TF32 speeds up the computation but introduces numeric errors that can cause small random fluctuations, e.g. in the presence of padding. 
In the following, we experiment both with and without TF32 (both at training and inference time) because padding has no effect on the final outputs only when TF32 is disabled.

\subsection{Impact of the Identified Bugs}
\label{subsec:impact_bug}

We evaluate the impact of the identified bugs (\S\ref{subsec:analysis}) on ASR and ST results by varying the IBSs as increasing the batch size introduces more padding, amplifying the effects of the bugs. Initially, experiments are conducted on 
our correct codebase (\smallcorrect). Subsequently, we enable single precision (TF32).
Then, we reintroduce the bugs individually (\bugone, \bugtwo, and \bugthree), and all together (\bugall).

\begin{table}[!tb]
\small
\setlength{\tabcolsep}{12pt}
    \centering
    \footnotesize
    \begin{tabular}{l|ccc}
    \specialrule{.1em}{.05em}{.05em} 
        \multirow{2}{*}{\textbf{Code}} & 
         \multicolumn{3}{c}{\textbf{IBS}} \\
        \cline{2-4}
         & 1 & 10 & 100 \\
        \specialrule{.1em}{.05em}{.05em} 
        \smallcorrect & 10.52 & 10.52 & 10.52 \\
        \hline
        + TF32 & 10.73 & 10.73 & 10.73 \\
        \quad + \bugone & 10.72 & 11.25* & 19.50* \\
        \quad + \bugtwo & 10.73 & 10.74 & 10.74 \\
        \quad + \bugthree & \textbf{10.46} & 10.62 & 10.73 \\
        \quad + \smallbug\textsubscript{1,2,3} & 11.32* & 14.25* & 54.56* \\
    \specialrule{.1em}{.05em}{.05em} 
    \end{tabular}
    \caption{WER
    for ASR with TF32 and bugs as IBS varies (1, 10, and 100 sentences). *~indicates that the difference with \smallcorrect{} is statistically significant.
    }
    \label{tab:ablationASR}
\end{table}

\paragraph{ASR}

Table \ref{tab:ablationASR} shows, in comparison to \smallcorrect{}, the impact of TF32 and of the different bugs on ASR performance.
First, TF32 causes a not statistically significant quality drop (+0.21 WER), which does not vary with the IBS (despite the presence of  minor variations in the outputs attested by  a slightly different number of generated words).
When the bugs are present (\bugone, \bugtwo, \bugthree), instead, the performance becomes sensitive to the IBS. This is particularly evident with \bugone, which significantly increases the error rate (+8.78 WER) when we introduce a considerable amount of padding (IBS=100).
It is noteworthy that most of the differences compared to the bug-free version (\smallcorrect{}) are not statistically significant, and the best result is achieved with \bugthree{} and 1 as IBS.
Only the presence of all bugs \bugall{} causes consistent and statistically significant quality drops. Nonetheless, the results with 1 and 10 as IBS are still far better than those obtained with Transformer architectures on the same benchmark (i.e. 26.61 by \citealt{CATTONI2021101155} and 15.6 by \citealt{gaido-etal-2021-ctc}). Moreover, their reproducibility is not hindered by the presence of bugs, as setting the IBS to any particular value consistently yields the same score. We can conclude that \textbf{\textit{even flawed code can produce competitive and reproducible results}} and, therefore, focusing only on these two aspects is not enough to ensure the trustworthiness of the code.

\begin{table}[!tb]
\small
\setlength{\tabcolsep}{2.5pt}
    \centering
    \footnotesize
    \begin{tabular}{l|ccc|ccc}
        \cline{2-7}
         & \multicolumn{3}{c|}{\textbf{en-de}} & \multicolumn{3}{c|}{\textbf{en-es}} \\
         \specialrule{.1em}{.05em}{.05em} 
        \multirow{2}{*}{\textbf{Code}} & \multicolumn{3}{c|}{\textbf{IBS}} & \multicolumn{3}{c}{\textbf{IBS}} \\
        \cline{2-7}
        & 1 & 10 & 100 & 1 & 10 & 100 \\
        \specialrule{.1em}{.05em}{.05em} 
        \smallcorrect & 24.67 & 24.67 & 24.67 & 30.34 & 30.34 & 30.34 \\
        \hline
        + TF32 & \textbf{24.84} & \textbf{24.84} & 24.83 & \textbf{30.63} & 30.62 & \textbf{30.63} \\
        \quad + \bugone & 24.52 & 24.65 & 24.67 & 29.53* & 29.41* & 27.71* \\
        \quad + \bugtwo & 24.56 & 24.57 & 24.58 & 30.53 & 30.53 & 30.53 \\
        \quad + \bugthree & 24.53 & 24.46 & 24.42 & 30.33 & 30.35 & 30.24\\
        \quad + \bugall & 24.68 & 24.58 & 23.23* & 28.57* & 27.81* & 21.15* \\
    \specialrule{.1em}{.05em}{.05em} 
    \end{tabular}
    \caption{BLEU
    for ST with TF32 and bugs as IBS varies (1, 10, and 100 sentences).
    *~indicates that the difference with \smallcorrect{} is statistically significant.
    }
    \label{tab:ablationST}
\end{table}

\begin{table}[t]
\setlength{\tabcolsep}{5.5pt}
\centering
\small
\begin{tabular}{l|cc}
\specialrule{.1em}{.05em}{.05em} 
\textbf{Model} & \textbf{en-de} & \textbf{en-es} 
 \\
\specialrule{.1em}{.05em}{.05em} 
 ESPNet \cite{inaguma-etal-2020-espnet} & 22.9 & 28.0 \\
 Fairseq \cite{wang2020fairseqs2t} & 22.7 & 27.2 \\
 Speechformer \cite{papi-etal-2021-speechformer} & 23.6 & 28.5 \\
 E2E + ML \citep{zhao-etal-2021-mutual} & - & 28.5 \\
 SATE (no KD) \citep{xu-etal-2021-stacked} & 24.1 & - \\
 E2E-ST-FS \citep{pmlr-v162-zhang22i} & 23.0 & 28.0 \\
 S2T-Perceiver \cite{Tsiamas-2023-Perceivers} & 24.2 & 28.0\\
 \hline
 Conformer \bugall & \textbf{24.7} & \textbf{28.6} \\
\specialrule{.1em}{.05em}{.05em} 
\end{tabular}
\caption{BLEU
of models trained on MuST-C en-de and en-es compared to Conformer \bugall{} (IBS=1).}
\label{tab:comparison_others}
\end{table}

\begin{table*}[t!]
\setcounter{table}{6}
\setlength{\tabcolsep}{5pt}
    \centering
    \small
    \begin{tabular}{c|c|c||c|c|c|c|c|c|c|c||c}
    \specialrule{.1em}{.05em}{.05em} 
        \textbf{Code} & \textbf{Model} & \textbf{IBS} & \textbf{en-de} & \textbf{en-es} & \textbf{en-fr} & \textbf{en-it} & \textbf{en-nl} & \textbf{en-pt} & \textbf{en-ro} & \textbf{en-ru} & \textbf{Avg} \\
        \specialrule{.1em}{.05em}{.05em} 
        \multirow{6}{*}{\correct} & \multirow{3}{*}{Conformer} & 1 & \multirow{3}{*}{24.67} & \multirow{3}{*}{30.34} & \multirow{3}{*}{36.22} & \multirow{3}{*}{25.73} & \multirow{3}{*}{30.04} & \multirow{3}{*}{\textbf{30.55}} & \multirow{3}{*}{23.43} & \multirow{3}{*}{17.29} & \multirow{3}{*}{27.28} \\
        & & 10 & & & & & & & & \\
        & & 100 & & & & & & & & \\
        \cline{2-12}
        & \multirow{3}{*}{\shortstack{Conformer\\+\\CTC Compr.}} & 1 & \multirow{3}{*}{24.97} & \multirow{3}{*}{30.48} & \multirow{3}{*}{\textbf{36.43}} & \multirow{3}{*}{\textbf{26.25*}} & \multirow{3}{*}{\textbf{30.31}} & \multirow{3}{*}{30.09\textsuperscript{$\dagger$}} & \multirow{3}{*}{\textbf{24.67*}} & \multirow{3}{*}{\textbf{17.35}} & \multirow{3}{*}{\textbf{27.57}} \\
        & & 10 & & & & & & & & \\
        & & 100 & & & & & & & & \\
        \hline
        \multirow{6}{*}{\bug\textsubscript{1,2,3}} & \multirow{3}{*}{Conformer} & 1 & 24.68 & 28.57 & 35.70 & 25.81 & 29.68 & 30.22 & 23.52 & 15.83 & 26.75 \\
        & & 10 & 24.58 & 27.81 & 35.65 & 25.70 & 29.35 & 30.02 & 23.43 & 15.36 & 26.49 \\
        & & 100 & 23.23 & 21.15 & 31.70 & 23.42 & 24.92 & 27.72 & 22.68 & 11.05 & 23.23 \\
        \cline{2-12}
        & \multirow{3}{*}{\shortstack{Conformer\\+\\CTC Compr.}} & 1 & 24.95 & 30.49* & 36.27* & 25.84 & 29.42 & 30.04 & 23.96* & 17.05* & 27.25 \\
        & & 10 & 25.21* & \textbf{30.72}* & 36.18* & 26.01 & 29.64 & 30.14 & 23.95* & 17.06* & 27.36 \\
        & & 100 & \textbf{25.26}* & 30.52* & 36.36* & 25.88* & 29.66* & 30.16* & 23.92* & 16.87* & 27.33 \\
    \specialrule{.1em}{.05em}{.05em} 
    \end{tabular}
    \caption{BLEU
    for ST of the correct/incorrect codebase with and without CTC 
    Compr. as IBS varies (1, 10, and 100).
    */\textsuperscript{$\dagger$} indicate that the improvement/degradation of CTC 
    Compr. is statistically significant.}
    \label{tab:ST}
\end{table*}

\paragraph{ST} Table \ref{tab:ablationST} reports the same study on the two most used sections of MuST-C (en-de, and en-es).
The behavior is quite different between the two, but the best scores are always obtained with TF32 and without bugs.
On en-es, \bugone{} causes statistically significant drops that increase 
with the IBS and are exacerbated if combined with the other two bugs (\bugall).
On en-de, instead, none of the bugs significantly impacts the results.
Interestingly, the result obtained with all bugs (\bugall) and 1 as IBS is slightly higher (+0.01) than that without bugs (\smallcorrect).
Furthermore, by comparing the scores obtained with all bugs (\bugall) and 1 as IBS with those of previous ST works (Table \ref{tab:comparison_others}), we can notice that, as previously observed for ASR, the presence of bugs is not evident from the results, which are still competitive with those of other models.
This supports our conclusion that \textbf{\textit{good (and reproducible) results do not imply code correctness}}, as this statement holds for different tasks and language pairs.

\subsection{Impact of Building on Incorrect Code}
\label{subsec:impact_code}

We now showcase how incorrect code can lead to misleading conclusions when experimenting with 
\mg{a}
new technique.
We choose to evaluate the CTC compression \citep{liu2020bridging,gaido-etal-2021-ctc} \mg{because we speculate that it limits the negative effects of the bugs identified in \S\ref{subsec:analysis}, as it reduces the sequence lengths and, in turn, the amount of padding.}
Introduced in the context of Transformer-based models, 
CTC compression \mg{reduces training and inference times, as well as VRAM requirements, while yielding}
minimal (not statistically significant) gains in terms of translation quality \cite{gaido-etal-2021-ctc}.
\mg{A detailed description of the CTC compression is provided in Appendix \ref{app:ctc_compr}.}


\begin{table}[h]
\setlength{\tabcolsep}{4.5pt}
\setcounter{table}{5}
    \centering
    \small
    \begin{tabular}{l|c|ccc}
    \specialrule{.1em}{.05em}{.05em} 
        \multirow{2}{*}{\textbf{Model}} & \multirow{2}{*}{\textbf{Code}} & \multicolumn{3}{c}{\textbf{IBS}} \\
        \cline{3-5}
         & & 1 & 10 & 100 \\
        \specialrule{.1em}{.05em}{.05em} 
        Conformer & \multirow{2}{*}{\smallcorrect} & 10.52 & 10.52 & 10.52 \\
        \hspace{0.5em}+ CTC Compr. & & 10.64 & 10.64 & 10.64 \\
        \hline
        Conformer & \multirow{2}{*}{\bugall} & 11.32 & 14.25 & 54.56 \\
        \hspace{0.5em}+ CTC Compr. & & 10.39* & \textbf{10.34}* & 10.81* \\ 
    \specialrule{.1em}{.05em}{.05em} 
    \end{tabular}
    \caption{WER
    for ASR of the correct/incorrect codebase with and without CTC
    Compr. as IBS varies (1, 10, and 100).
    * indicates that the improvement of CTC 
    Compr. is statistically significant.}
    \label{tab:ASR}
\end{table}

\paragraph{ASR}
Table \ref{tab:ASR} shows the effects on 
ASR performance of 
introducing CTC compression (CTC Compr.)
into the codebase with all bugs~(\bugall{}) and without them~(\smallcorrect).
\mg{CTC} compression causes a small and not statistically significant performance degradation (+0.12 WER) when the correct implementation (\smallcorrect{}) is used (in accordance with the findings
\mg{on}
the Transformer architecture).
When bugs are present in the codebase (\bugall), instead, the 
outcome is overturned:
CTC compression brings statistically significant
gains
(-0.93
WER even with 1 as IBS).
These observations lead to the conclusion that \textit{\textbf{building on incorrect code can produce misleading findings}}.
Besides, the best overall result is achieved with the \bugall{} codebase (with 10 as IBS and CTC compression), reiterating that high scores do not imply code correctness.

\paragraph{ST}
Table \ref{tab:ST} reports the same analysis on the 8 language pairs of MuST-C.
As in ASR, the presence of bugs (\bugall) unduly rewards the CTC compression mechanism, which yields statistically significant gains on all the languages with 100 as IBS and on 4/5 out of 8 languages with 1/10 as IBS.
With the bug-free version (\smallcorrect), instead, the improvements are statistically significant only on two language pairs (en-it, and en-ro), while on en-pt there is a statistically significant degradation.
On average over all language pairs, the gain brought by 
CTC compression in the presence of bugs (\bugall) ranges from 0.5 BLEU (with 1 as IBS) to 4.1 BLEU (with 100 as IBS), while it is only of 0.29 BLEU with the correct code~(\smallcorrect). 
We can hence confirm that \textbf{\textit{the presence of bugs
leads to erroneous findings}} also in the ST task, as
CTC compression seems to significantly improve translation quality, which is not the case with the correct Conformer implementation
(as well as with Transformer, as proved by \citealt{gaido-etal-2021-ctc}).
Moreover, the best scores for en-de and en-es are achieved with the presence of all bugs (\bugall), and the average performance gap between codebases with (\bugall{}) and without (\smallcorrect{}) bugs can be as little as 0.21 BLEU (when IBS is 10) and may be further
narrowed, or even overturned, 
by \enquote{tuning} the IBS.
This demonstrates again the \textit{\textbf{impossibility to assess code correctness only by looking at the results}}.

%% file: sections/proposal.tex
After demonstrating
that the current tendency to assess code correctness solely based on the reported results (\S\ref{sec:core-idea}) potentially leads to wrong findings (\S\ref{sec:case-study}), in this section we propose countermeasures. Specifically, we aim at 
fostering the adoption of SQA best practices in two ways: \textit{1)} by releasing a  Python package
(\texttt{pangoliNN}) for testing neural networks and assisting researchers in the verification of code correctness (\S\ref{subsec:pangolinn});  \textit{2)} by proposing the integration of current conference checklists with recommendations for SQA best practices (\S\ref{subsec:checklist}).

\subsection{\texttt{pangoliNN}}
\label{subsec:pangolinn}
As discussed in \S\ref{sec:sqa}, testing software is the only way to enforce that it works correctly. Therefore, as the first and foremost method to increase code correctness, we recommend the extensive implementation and adoption of Unit Tests (UTs) to check that the code has the expected behavior \citep{10.1145/987305.987309,10.1145/800027.808473,10.5555/1349795,8048665}.
Ideally, this should be done prior to writing the actual code, following the so-called ``test-driven development practice'' \citep{beck2002driven}.
UTs should cover all the assumptions about how the code works (e.g., ensuring that the presence of padding does not alter the results). While achieving complete test coverage is a utopian objective, the higher the coverage, the higher the quality of the codebase.

To ease this work, we introduce \texttt{pangoliNN},\footnote{As a pangolin looks for bugs and catches them, this library aims at finding bugs in neural networks (NN). Hence the name.} a Python package specifically designed for testing neural modules. Built upon the widely used PyTorch library \citep{Paszke-2019-pytorch}, \texttt{pangoliNN} offers a collection of pre-defined tests that enforce specific behaviors of the modules.\footnote{\label{foot:suppl_mat}\mg{See \url{https://readthedocs.org/projects/pangolinn/}.}}
Its primary objective is to simplify and expedite the process of testing neural networks, alleviating researchers from the burden of creating UTs from scratch.
Indeed, writing UTs may initially be perceived as
an additional and undesirable cost, although ample literature dispels this perception. \citet{Williams-2003-TDD}, for instance, proved that the inclusion of UTs does not hamper code-writing productivity, 
\citet{oro3667,Ellims2006} showed that the perceived cost is \enquote{\textit{exaggerated}}, and \citet{google-ut}
that the initial overhead\footnote{Estimated in 16\%-35\% of the overall software development cost \citep{George-2003-tdd,google-ut}.} pays off by saving time spent on manual experiments.


Furthermore, unlike manual experiments that are often resource-intensive and environmentally impactful \citep{strubell-etal-2019-energy}, UTs are generally lightweight (e.g., they do not involve any training phase). As such, writing UTs would contribute to the environmental sustainability of NLP research,
facilitating the transition to Green AI~\citep{greenai}. Also, UTs do not require any pre-trained model to run, as their nature and goal greatly differ from assessing the quality of a trained model, as recently proposed with behavioral testing \citep{ribeiro-etal-2020-beyond}.
Through behavioral testing, we check whether a specific instance properly handles different aspects, such as linguistic phenomena (e.g., negation, co-references), and/or produces correct outputs with challenging inputs. 
Through UTs, instead, we assess the behavior and robustness of the code itself (rather than of model instances), by verifying whether assumptions about properties of the network or about its behavior in specific conditions (e.g., not being influenced by the presence of padding) are respected.

Currently, \texttt{pangoliNN} includes tests for two aspects: \textit{i)} proper handling of padding and batching, ensuring consistent output of neural modules irrespective of padding presence; and \textit{ii)} addressing causality by verifying the independence of module output from future elements in the input sequence, which is crucial for autoregressive and other sequence-to-sequence models.
It also features comprehensive documentation\textsuperscript{\ref{foot:suppl_mat}} with simple examples to guide researchers in its usage. Moreover, \texttt{pangoliNN} itself is extensively unit tested, and these UTs provide additional implicit guidance on how to effectively utilize the package.

Despite being in its initial stage (for the limited number of tests currently covered), we argue that the first release of \texttt{pangoliNN} represents a milestone toward increasing the quality and trustworthiness of NLP research code and, in turn, outcomes.
We hope that it will be embraced and expanded upon by the research community, with the integration of additional tests, ultimately growing it into a comprehensive testing library for neural networks.

\subsection{Code-quality Checklist}
\label{subsec:checklist}

As a complementary initiative, we also propose to integrate the existing conference checklists with questions targeting the improvement of code correctness and quality in research (Table \ref{tab:checklist}). 
It is worth mentioning that we have strictly adhered to these guidelines throughout
the development of both \texttt{pangoliNN} and of our padding-safe implementation of the Conformer architecture.

The first questions (\textbf{Q1-2}) focus on the adoption of UTs (possibly leveraging \texttt{pangoliNN}), whose importance has been stressed
in the previous section.
However, the presence of UTs alone does not guarantee that the code works.
Indeed, UTs should be executed every time the codebase is modified, even in case of a seemingly \textit{unrelated change}, as the validity of a test expires whenever the software is edited (as seen in \S\ref{sec:sqa}).
This is commonly enforced through continuous integration (CI), which executes all UTs at every code change~\citep{Duvall-2007-ci}. 
A running and successful CI  offers the supplementary advantage of providing implicit guidance on the installation and execution of the code for individuals attempting to replicate a study.
Furthermore, it mitigates the occurrence of replication failures due to syntax or runtime errors, as it often happens in current NLP artifacts \citep{arvan-etal-2022-reproducibility-code}.
Such failures can arise because the released version of the code might slightly differ from those used for the experiments due to small refactorings prior to the release.
For this reason, \textbf{Q3} and \textbf{Q4} respectively focus on test execution and on the presence of a CI, so as to ensure that the checks of the UTs are actually respected.

\begin{table}[!tb]
\setcounter{table}{7}
\begin{tcbitemize}[%
        raster columns=1,
        raster equal height,
        raster width=.47\textwidth,
        before=,after=\hfill,
        boxsep=3pt, left=4pt, right=6pt, 
        colframe=teal!75!black,colback=white,
        fonttitle=\large\bfseries,
        halign=left,
        ]
\tcbitem
\begin{enumerate}[itemsep=1pt,leftmargin=*]
\small
\justifying
\item \textsf{Have you tested your code with relevant tests? 
}
\item \textsf{Have you tested assumptions about code behavior with Unit Tests (UTs)?}
\item \textsf{Have UTs been executed on the code version used for the experiments and, if applicable, the publicly released version?}
\item \textsf{Does the repository contain a continuous integration that 
runs the UTs?}
\item \textsf{Has every contribution to the codebase been reviewed by at least one person?}
\end{enumerate}
\end{tcbitemize}
\caption{Code-quality Checklist.}
\label{tab:checklist}
\end{table}

Lastly, we encourage (\textbf{Q5}) the adoption of a code reviewing practice \citep{7589787}, in which all changes are reviewed and approved by a person different from the code author.\footnote{The reviewer(s) can be any person
with basic knowledge of the codebase, such as lab teammates or advisors.}
Code review is a lightweight and informal process compared to code inspection \citep{fagan-1976} and has been shown to cause little overhead for most code changes \citep{10.1145/3183519.3183525}.
It consists in reading and commenting on the source code for a change, which should be kept small and focused on one single aspect or new feature.
This aims not only at avoiding bugs, but also at improving code readability and documentation \citep{10.1007/978-3-319-69926-4_9,Chen2019,Bahaidarah-2022-toward,Trisovic2022}, and, in turn, reusability and reproducibility.
In addition, it serves as a powerful tool for knowledge transfer \citep{10.5555/2486788.2486882}, thus novices would particularly benefit from it.

As a final note, we would like to emphasize that our proposed \enquote{Code-quality Checklist} should be interpreted as the \enquote{Reproducibility Checklist} now required by many top-tier venues: though strongly encouraged, following
the checklist is not mandatory for paper submissions and its intent is fostering software quality and correctness rather than certifying it. Specifically, it will encourage awareness of SQA concepts and coding best practices, especially among researchers who have not been exposed to them during their education.

%% file: acl_latex.bbl
\begin{thebibliography}{102}
\expandafter\ifx\csname natexlab\endcsname\relax\def\natexlab#1{#1}\fi

\bibitem[{Arvan et~al.(2022)Arvan, Pina, and Parde}]{arvan-etal-2022-reproducibility-code}
Mohammad Arvan, Lu{\'\i}s Pina, and Natalie Parde. 2022.
\newblock \href {https://aclanthology.org/2022.emnlp-main.150} {Reproducibility in computational linguistics: Is source code enough?}
\newblock In \emph{Proceedings of the 2022 Conference on Empirical Methods in Natural Language Processing}, pages 2350--2361, Abu Dhabi, United Arab Emirates.

\bibitem[{Bacchelli and Bird(2013)}]{10.5555/2486788.2486882}
Alberto Bacchelli and Christian Bird. 2013.
\newblock Expectations, outcomes, and challenges of modern code review.
\newblock In \emph{Proceedings of the 2013 International Conference on Software Engineering}, ICSE '13, page 712–721. IEEE Press.

\bibitem[{Bahaidarah et~al.(2022)Bahaidarah, Hung, De~Melo~Oliveira, Penumaka, Rosario, and Trisovic}]{Bahaidarah-2022-toward}
Layan Bahaidarah, Ethan Hung, Andreas~F. De~Melo~Oliveira, Jyotsna Penumaka, Lukas Rosario, and Ana Trisovic. 2022.
\newblock \href {https://doi.org/10.1109/eScience55777.2022.00079} {Toward reusable science with readable code and reproducibility}.
\newblock In \emph{2022 IEEE 18th International Conference on e-Science (e-Science)}, pages 437--439, Los Alamitos, CA, USA.

\bibitem[{Baker(2016)}]{Baker2016}
Monya Baker. 2016.
\newblock \href {https://doi.org/10.1038/533452a} {1,500 scientists lift the lid on reproducibility}.
\newblock \emph{Nature}, 533(7604):452--454.

\bibitem[{Barba(2019)}]{barba-2019-praxis}
Lorena~A. Barba. 2019.
\newblock \href {https://doi.org/10.1109/MCSE.2018.2881905} {Praxis of reproducible computational science}.
\newblock \emph{Computing in Science \& Engineering}, 21(1):73--78.

\bibitem[{Baum et~al.(2017)Baum, Le{\ss}mann, and Schneider}]{10.1007/978-3-319-69926-4_9}
Tobias Baum, Hendrik Le{\ss}mann, and Kurt Schneider. 2017.
\newblock {The Choice of Code Review Process: A Survey on the State of the Practice}.
\newblock In \emph{Product-Focused Software Process Improvement}, pages 111--127, Cham. Springer International Publishing.

\bibitem[{Baum et~al.(2016)Baum, Liskin, Niklas, and Schneider}]{7589787}
Tobias Baum, Olga Liskin, Kai Niklas, and Kurt Schneider. 2016.
\newblock \href {https://doi.org/10.1109/QRS.2016.19} {A faceted classification scheme for change-based industrial code review processes}.
\newblock In \emph{2016 IEEE International Conference on Software Quality, Reliability and Security (QRS)}, pages 74--85.

\bibitem[{Beck(2002)}]{beck2002driven}
Kent Beck. 2002.
\newblock \emph{Test Driven Development. By Example (Addison-Wesley Signature)}.
\newblock Addison-Wesley Longman, Amsterdam.

\bibitem[{Belz(2022)}]{10.1162/coli_a_00448}
Anya Belz. 2022.
\newblock \href {https://doi.org/10.1162/coli_a_00448} {{{A Metrological Perspective on Reproducibility in NLP}}}.
\newblock \emph{Computational Linguistics}, 48(4):1125--1135.

\bibitem[{Belz et~al.(2021{\natexlab{a}})Belz, Agarwal, Shimorina, and Reiter}]{belz-etal-2021-systematic}
Anya Belz, Shubham Agarwal, Anastasia Shimorina, and Ehud Reiter. 2021{\natexlab{a}}.
\newblock \href {https://doi.org/10.18653/v1/2021.eacl-main.29} {A systematic review of reproducibility research in natural language processing}.
\newblock In \emph{Proceedings of the 16th Conference of the European Chapter of the Association for Computational Linguistics: Main Volume}, pages 381--393, Online. Association for Computational Linguistics.

\bibitem[{Belz et~al.(2022)Belz, Popovic, and Mille}]{belz-etal-2022-quantified}
Anya Belz, Maja Popovic, and Simon Mille. 2022.
\newblock \href {https://doi.org/10.18653/v1/2022.acl-long.2} {Quantified reproducibility assessment of {NLP} results}.
\newblock In \emph{Proceedings of the 60th Annual Meeting of the Association for Computational Linguistics (Volume 1: Long Papers)}, pages 16--28, Dublin, Ireland.

\bibitem[{Belz et~al.(2021{\natexlab{b}})Belz, Shimorina, Agarwal, and Reiter}]{belz-etal-2021-reprogen}
Anya Belz, Anastasia Shimorina, Shubham Agarwal, and Ehud Reiter. 2021{\natexlab{b}}.
\newblock \href {https://aclanthology.org/2021.inlg-1.24} {The {R}epro{G}en shared task on reproducibility of human evaluations in {NLG}: Overview and results}.
\newblock In \emph{Proceedings of the 14th International Conference on Natural Language Generation}, pages 249--258, Aberdeen, Scotland, UK. Association for Computational Linguistics.

\bibitem[{Belz et~al.(2023)Belz, Thomson, Reiter, and Mille}]{belz-etal-2023-non}
Anya Belz, Craig Thomson, Ehud Reiter, and Simon Mille. 2023.
\newblock \href {https://doi.org/10.18653/v1/2023.findings-acl.226} {{Non-Repeatable Experiments and Non-Reproducible Results: The Reproducibility Crisis in Human Evaluation in {NLP}}}.
\newblock In \emph{Findings of the Association for Computational Linguistics: ACL 2023}, pages 3676--3687, Toronto, Canada. Association for Computational Linguistics.

\bibitem[{B{\'e}rard et~al.(2016)B{\'e}rard, Pietquin, Servan, and Besacier}]{berard_2016}
Alexandre B{\'e}rard, Olivier Pietquin, Christophe Servan, and Laurent Besacier. 2016.
\newblock {Listen and Translate: A Proof of Concept for End-to-End Speech-to-Text Translation}.
\newblock In \emph{NIPS Workshop on end-to-end learning for speech and audio processing}, Barcelona, Spain.

\bibitem[{Branco et~al.(2020)Branco, Calzolari, Vossen, Van~Noord, van Uytvanck, Silva, Gomes, Moreira, and Elbers}]{branco-etal-2020-shared}
Ant{\'o}nio Branco, Nicoletta Calzolari, Piek Vossen, Gertjan Van~Noord, Dieter van Uytvanck, Jo{\~a}o Silva, Lu{\'\i}s Gomes, Andr{\'e} Moreira, and Willem Elbers. 2020.
\newblock \href {https://aclanthology.org/2020.lrec-1.680} {{A Shared Task of a New, Collaborative Type to Foster Reproducibility: A First Exercise in the Area of Language Science and Technology with {REPROLANG}2020}}.
\newblock In \emph{Proceedings of the Twelfth Language Resources and Evaluation Conference}, pages 5539--5545, Marseille, France. European Language Resources Association.

\bibitem[{Buckley and Poston(1984)}]{Buckley-1984-sqa}
Fletcher~J. Buckley and Robert Poston. 1984.
\newblock \href {https://doi.org/10.1109/TSE.1984.5010196} {Software quality assurance}.
\newblock \emph{IEEE Transactions on Software Engineering}, SE-10(1):36--41.

\bibitem[{Cattoni et~al.(2021)Cattoni, {Di Gangi}, Bentivogli, Negri, and Turchi}]{CATTONI2021101155}
Roldano Cattoni, Mattia~Antonino {Di Gangi}, Luisa Bentivogli, Matteo Negri, and Marco Turchi. 2021.
\newblock \href {https://doi.org/https://doi.org/10.1016/j.csl.2020.101155} {Must-c: A multilingual corpus for end-to-end speech translation}.
\newblock \emph{Computer Speech \& Language}, 66:101155.

\bibitem[{Chen et~al.(2019)Chen, Dallmeier-Tiessen, Dasler, Feger, Fokianos, Gonzalez, Hirvonsalo, Kousidis, Lavasa, Mele, Rodriguez, {\v{S}}imko, Smith, Trisovic, Trzcinska, Tsanaktsidis, Zimmermann, Cranmer, Heinrich, Watts, Hildreth, Lloret~Iglesias, Lassila-Perini, and Neubert}]{Chen2019}
Xiaoli Chen, S{\"u}nje Dallmeier-Tiessen, Robin Dasler, Sebastian Feger, Pamfilos Fokianos, Jose~Benito Gonzalez, Harri Hirvonsalo, Dinos Kousidis, Artemis Lavasa, Salvatore Mele, Diego~Rodriguez Rodriguez, Tibor {\v{S}}imko, Tim Smith, Ana Trisovic, Anna Trzcinska, Ioannis Tsanaktsidis, Markus Zimmermann, Kyle Cranmer, Lukas Heinrich, Gordon Watts, Michael Hildreth, Lara Lloret~Iglesias, Kati Lassila-Perini, and Sebastian Neubert. 2019.
\newblock \href {https://doi.org/10.1038/s41567-018-0342-2} {Open is not enough}.
\newblock \emph{Nature Physics}, 15(2):113--119.

\bibitem[{Chorowski et~al.(2014)Chorowski, Bahdanau, Cho, and Bengio}]{Chorowski-2014-asr}
Jan Chorowski, Dzmitry Bahdanau, Kyunghyun Cho, and Yoshua Bengio. 2014.
\newblock End-to-end continuous speech recognition using attention-based recurrent nn: First results.
\newblock In \emph{NIPS 2014 Workshop on Deep Learning, December 2014}.

\bibitem[{Dai et~al.(2019)Dai, Yang, Yang, Carbonell, Le, and Salakhutdinov}]{dai-etal-2019-transformer}
Zihang Dai, Zhilin Yang, Yiming Yang, Jaime Carbonell, Quoc Le, and Ruslan Salakhutdinov. 2019.
\newblock \href {https://doi.org/10.18653/v1/P19-1285} {Transformer-{XL}: Attentive language models beyond a fixed-length context}.
\newblock In \emph{Proceedings of the 57th Annual Meeting of the Association for Computational Linguistics}, pages 2978--2988, Florence, Italy. Association for Computational Linguistics.

\bibitem[{Dauphin et~al.(2017)Dauphin, Fan, Auli, and Grangier}]{Dauphin-2017-glu}
Yann~N. Dauphin, Angela Fan, Michael Auli, and David Grangier. 2017.
\newblock Language modeling with gated convolutional networks.
\newblock In \emph{Proceedings of the 34th International Conference on Machine Learning - Volume 70}, ICML'17, page 933–941. JMLR.org.

\bibitem[{Denkowski and Neubig(2017)}]{denkowski-neubig-2017-stronger}
Michael Denkowski and Graham Neubig. 2017.
\newblock \href {https://doi.org/10.18653/v1/W17-3203} {Stronger baselines for trustable results in neural machine translation}.
\newblock In \emph{Proceedings of the First Workshop on Neural Machine Translation}, pages 18--27, Vancouver. Association for Computational Linguistics.

\bibitem[{Deutsch and Willis(1988)}]{10.5555/48813}
Michael~S. Deutsch and Ronald~R. Willis. 1988.
\newblock \emph{Software Quality Engineering: A Total Technical and Management Approach}.
\newblock Prentice-Hall, Inc., USA.

\bibitem[{Di~Gangi et~al.(2020)Di~Gangi, Gaido, Negri, and Turchi}]{di-gangi-etal-2020-target}
Mattia~A. Di~Gangi, Marco Gaido, Matteo Negri, and Marco Turchi. 2020.
\newblock \href {https://aclanthology.org/2020.amta-research.13} {On target segmentation for direct speech translation}.
\newblock In \emph{Proceedings of the 14th Conference of the Association for Machine Translation in the Americas (Volume 1: Research Track)}, pages 137--150, Virtual. Association for Machine Translation in the Americas.

\bibitem[{Di~Gangi et~al.(2019)Di~Gangi, Negri, and Turchi}]{gangi19_interspeech}
Mattia~A. Di~Gangi, Matteo Negri, and Marco Turchi. 2019.
\newblock \href {https://doi.org/10.21437/Interspeech.2019-3045} {{Adapting Transformer to End-to-End Spoken Language Translation}}.
\newblock In \emph{Proc. Interspeech 2019}, pages 1133--1137.

\bibitem[{Dodge et~al.(2019)Dodge, Gururangan, Card, Schwartz, and Smith}]{dodge-etal-2019-show}
Jesse Dodge, Suchin Gururangan, Dallas Card, Roy Schwartz, and Noah~A. Smith. 2019.
\newblock \href {https://doi.org/10.18653/v1/D19-1224} {Show your work: Improved reporting of experimental results}.
\newblock In \emph{Proceedings of the 2019 Conference on Empirical Methods in Natural Language Processing and the 9th International Joint Conference on Natural Language Processing (EMNLP-IJCNLP)}, pages 2185--2194, Hong Kong, China. Association for Computational Linguistics.

\bibitem[{Dong et~al.(2018)Dong, Xu, and Xu}]{8462506}
Linhao Dong, Shuang Xu, and Bo~Xu. 2018.
\newblock \href {https://doi.org/10.1109/ICASSP.2018.8462506} {Speech-transformer: A no-recurrence sequence-to-sequence model for speech recognition}.
\newblock In \emph{2018 IEEE International Conference on Acoustics, Speech and Signal Processing (ICASSP)}, pages 5884--5888.

\bibitem[{Dror et~al.(2018)Dror, Baumer, Shlomov, and Reichart}]{dror-etal-2018-hitchhikers}
Rotem Dror, Gili Baumer, Segev Shlomov, and Roi Reichart. 2018.
\newblock \href {https://doi.org/10.18653/v1/P18-1128} {The hitchhiker{'}s guide to testing statistical significance in natural language processing}.
\newblock In \emph{Proceedings of the 56th Annual Meeting of the Association for Computational Linguistics (Volume 1: Long Papers)}, pages 1383--1392, Melbourne, Australia. Association for Computational Linguistics.

\bibitem[{Duvall et~al.(2007)Duvall, Matyas, and Glovert}]{Duvall-2007-ci}
Paul~M. Duvall, Steve Matyas, and Andrew Glovert. 2007.
\newblock \emph{Continuous Integration: Improving Software Quality and Reducing Risk}.
\newblock Addison-Wesley Professional.

\bibitem[{Ellims et~al.(2004)Ellims, Bridges, and Ince}]{oro3667}
Michael Ellims, James Bridges, and Darrel~C. Ince. 2004.
\newblock \href {http://oro.open.ac.uk/3667/} {Unit testing in practice}.
\newblock In \emph{Proceedings of the 15th International Symposium on Software Reliability Engineering (ISSRE?04)}. IEEE.

\bibitem[{Ellims et~al.(2006)Ellims, Bridges, and Ince}]{Ellims2006}
Michael Ellims, James Bridges, and Darrel~C. Ince. 2006.
\newblock \href {https://doi.org/10.1007/s10664-006-5964-9} {The economics of unit testing}.
\newblock \emph{Empirical Software Engineering}, 11(1):5--31.

\bibitem[{Fagan(1976)}]{fagan-1976}
M.~E. Fagan. 1976.
\newblock \href {https://doi.org/10.1147/sj.153.0182} {Design and code inspections to reduce errors in program development}.
\newblock \emph{IBM Systems Journal}, 15(3):182--211.

\bibitem[{Gaido et~al.(2021)Gaido, Cettolo, Negri, and Turchi}]{gaido-etal-2021-ctc}
Marco Gaido, Mauro Cettolo, Matteo Negri, and Marco Turchi. 2021.
\newblock \href {https://doi.org/10.18653/v1/2021.eacl-main.57} {{CTC}-based compression for direct speech translation}.
\newblock In \emph{Proceedings of the 16th Conference of the European Chapter of the Association for Computational Linguistics: Main Volume}, pages 690--696, Online.

\bibitem[{Gaido et~al.(2022)Gaido, Papi, Fucci, Fiameni, Negri, and Turchi}]{gaido-etal-2022-efficient}
Marco Gaido, Sara Papi, Dennis Fucci, Giuseppe Fiameni, Matteo Negri, and Marco Turchi. 2022.
\newblock \href {https://doi.org/10.18653/v1/2022.iwslt-1.13} {Efficient yet competitive speech translation: {FBK}@{IWSLT}2022}.
\newblock In \emph{Proceedings of the 19th International Conference on Spoken Language Translation (IWSLT 2022)}, pages 177--189, Dublin, Ireland (in-person and online). Association for Computational Linguistics.

\bibitem[{Garlan and Shaw(1993)}]{intro_soft_architectures}
David Garlan and Mary Shaw. 1993.
\newblock \href {https://doi.org/10.1142/9789812798039_0001} {\emph{AN INTRODUCTION TO SOFTWARE ARCHITECTURE}}, pages 1--39.

\bibitem[{Gehrmann et~al.(2022)Gehrmann, Clark, and Sellam}]{gehrmann2022repairing}
Sebastian Gehrmann, Elizabeth Clark, and Thibault Sellam. 2022.
\newblock Repairing the cracked foundation: A survey of obstacles in evaluation practices for generated text.
\newblock \emph{arXiv preprint arXiv:2202.06935}.

\bibitem[{George and Williams(2003)}]{George-2003-tdd}
Boby George and Laurie Williams. 2003.
\newblock \href {https://doi.org/10.1145/952532.952753} {An initial investigation of test driven development in industry}.
\newblock In \emph{Proceedings of the 2003 ACM Symposium on Applied Computing}, SAC '03, page 1135–1139, New York, NY, USA. Association for Computing Machinery.

\bibitem[{Glass(1992)}]{10.5555/121025}
Robert~L. Glass. 1992.
\newblock \emph{Building Quality Software}.
\newblock Prentice-Hall, Inc., USA.

\bibitem[{Goodenough and Gerhart(1975)}]{10.1145/800027.808473}
John~B. Goodenough and Susan~L. Gerhart. 1975.
\newblock \href {https://doi.org/10.1145/800027.808473} {Toward a theory of test data selection}.
\newblock In \emph{Proceedings of the International Conference on Reliable Software}, page 493–510, New York, NY, USA.

\bibitem[{Graves et~al.(2006)Graves, Fern{\'a}ndez, Gomez, and Schmidhuber}]{Graves2006ConnectionistTC}
Alex Graves, Santiago Fern{\'a}ndez, Faustino~J. Gomez, and J{\"u}rgen Schmidhuber. 2006.
\newblock {Connectionist Temporal Classification: Labelling Unsegmented Sequence Data with Recurrent Neural Networks}.
\newblock In \emph{Proceedings of the 23rd international conference on Machine learning (ICML)}, pages 369--376, Pittsburgh, Pennsylvania.

\bibitem[{Graves and Jaitly(2014)}]{pmlr-v32-graves14}
Alex Graves and Navdeep Jaitly. 2014.
\newblock \href {https://proceedings.mlr.press/v32/graves14.html} {Towards end-to-end speech recognition with recurrent neural networks}.
\newblock In \emph{Proceedings of the 31st International Conference on Machine Learning}, volume~32 of \emph{Proceedings of Machine Learning Research}, pages 1764--1772, Bejing, China.

\bibitem[{Gulati et~al.(2020)Gulati, Qin, Chiu, Parmar, Zhang, Yu, Han, Wang, Zhang, Wu, and Pang}]{gulati20_interspeech}
Anmol Gulati, James Qin, Chung-Cheng Chiu, Niki Parmar, Yu~Zhang, Jiahui Yu, Wei Han, Shibo Wang, Zhengdong Zhang, Yonghui Wu, and Ruoming Pang. 2020.
\newblock \href {https://doi.org/10.21437/Interspeech.2020-3015} {{Conformer: Convolution-augmented Transformer for Speech Recognition}}.
\newblock In \emph{Proc. Interspeech 2020}, pages 5036--5040.

\bibitem[{Gundersen(2019)}]{Gundersen_2019}
Odd~E. Gundersen. 2019.
\newblock \href {https://doi.org/10.1609/aimag.v40i4.5185} {Standing on the feet of giants — reproducibility in ai}.
\newblock \emph{AI Magazine}, 40(4):9--23.

\bibitem[{Gundersen and Kjensmo(2018)}]{Gundersen_Kjensmo_2018}
Odd~E. Gundersen and Sigbjørn Kjensmo. 2018.
\newblock \href {https://doi.org/10.1609/aaai.v32i1.11503} {State of the art: Reproducibility in artificial intelligence}.
\newblock \emph{Proceedings of the AAAI Conference on Artificial Intelligence}, 32(1).

\bibitem[{Guo et~al.(2021)Guo, Boyer, Chang, Hayashi, Higuchi, Inaguma, Kamo, Li, Garcia-Romero, Shi, Shi, Watanabe, Wei, Zhang, and Zhang}]{conformer-sota}
Pengcheng Guo, Florian Boyer, Xuankai Chang, Tomoki Hayashi, Yosuke Higuchi, Hirofumi Inaguma, Naoyuki Kamo, Chenda Li, Daniel Garcia-Romero, Jiatong Shi, Jing Shi, Shinji Watanabe, Kun Wei, Wangyou Zhang, and Yuekai Zhang. 2021.
\newblock \href {https://doi.org/10.1109/ICASSP39728.2021.9414858} {Recent developments on espnet toolkit boosted by conformer}.
\newblock In \emph{ICASSP 2021 - 2021 IEEE International Conference on Acoustics, Speech and Signal Processing (ICASSP)}, pages 5874--5878.

\bibitem[{Hevery(2009)}]{google-ut}
Miško Hevery. 2009.
\newblock {Cost of Testing}.
\newblock \url{https://testing.googleblog.com/2009/10/cost-of-testing.html}.
\newblock Accessed: 2023-02-06.

\bibitem[{Huizinga and Kolawa(2007)}]{10.5555/1349795}
Dorota Huizinga and Adam Kolawa. 2007.
\newblock \emph{Automated Defect Prevention: Best Practices in Software Management}.
\newblock Wiley-IEEE Press.

\bibitem[{Inaguma et~al.(2021)Inaguma, Higuchi, Duh, Kawahara, and Watanabe}]{inaguma2021non}
Hirofumi Inaguma, Yosuke Higuchi, Kevin Duh, Tatsuya Kawahara, and Shinji Watanabe. 2021.
\newblock Non-autoregressive end-to-end speech translation with parallel autoregressive rescoring.
\newblock \emph{arXiv preprint arXiv:2109.04411}.

\bibitem[{Inaguma et~al.(2020)Inaguma, Kiyono, Duh, Karita, Yalta, Hayashi, and Watanabe}]{inaguma-etal-2020-espnet}
Hirofumi Inaguma, Shun Kiyono, Kevin Duh, Shigeki Karita, Nelson Yalta, Tomoki Hayashi, and Shinji Watanabe. 2020.
\newblock \href {https://www.aclweb.org/anthology/2020.acl-demos.34} {{ESP}net-{ST}: All-in-one speech translation toolkit}.
\newblock In \emph{Proceedings of the 58th Annual Meeting of the Association for Computational Linguistics: System Demonstrations}, pages 302--311, Online. Association for Computational Linguistics.

\bibitem[{Ioffe and Szegedy(2015)}]{ioffe-2015-batchnorm}
Sergey Ioffe and Christian Szegedy. 2015.
\newblock Batch normalization: Accelerating deep network training by reducing internal covariate shift.
\newblock In \emph{Proceedings of the 32nd International Conference on International Conference on Machine Learning - Volume 37}, ICML'15, page 448–456. JMLR.org.

\bibitem[{ISO/IEC(2001)}]{ISO9126}
ISO/IEC. 2001.
\newblock \emph{ISO/IEC 9126. Software engineering -- Product quality}.
\newblock ISO/IEC.

\bibitem[{ISO/IEC(2010)}]{ISO/IEC2010}
ISO/IEC. 2010.
\newblock \emph{ISO/IEC 25010 System and software quality models}.
\newblock ISO/IEC.

\bibitem[{Kassab et~al.(2017)Kassab, DeFranco, and Laplante}]{8048665}
Mohamad Kassab, Joanna~F. DeFranco, and Phillip~A. Laplante. 2017.
\newblock \href {https://doi.org/10.1109/MS.2017.3571582} {Software testing: The state of the practice}.
\newblock \emph{IEEE Software}, 34(5):46--52.

\bibitem[{Kingma and Ba(2015)}]{journals/corr/KingmaB14}
Diederik~P. Kingma and Jimmy Ba. 2015.
\newblock \href {http://arxiv.org/abs/1412.6980} {Adam: {A} method for stochastic optimization}.
\newblock In \emph{3rd International Conference on Learning Representations, {ICLR} 2015, San Diego, CA, USA, May 7-9, 2015, Conference Track Proceedings}.

\bibitem[{Koehn(2004)}]{koehn-2004-statistical}
Philipp Koehn. 2004.
\newblock \href {https://aclanthology.org/W04-3250} {Statistical significance tests for machine translation evaluation}.
\newblock In \emph{Proceedings of the 2004 Conference on Empirical Methods in Natural Language Processing}, pages 388--395, Barcelona, Spain. Association for Computational Linguistics.

\bibitem[{Kuchaiev et~al.(2019)Kuchaiev, Li, Nguyen, Hrinchuk, Leary, Ginsburg, Kriman, Beliaev, Lavrukhin, Cook et~al.}]{kuchaiev2019nemo}
Oleksii Kuchaiev, Jason Li, Huyen Nguyen, Oleksii Hrinchuk, Ryan Leary, Boris Ginsburg, Samuel Kriman, Stanislav Beliaev, Vitaly Lavrukhin, Jack Cook, et~al. 2019.
\newblock Nemo: a toolkit for building ai applications using neural modules.
\newblock \emph{arXiv preprint arXiv:1909.09577}.

\bibitem[{Kudo and Richardson(2018)}]{kudo-richardson-2018-sentencepiece}
Taku Kudo and John Richardson. 2018.
\newblock \href {https://doi.org/10.18653/v1/D18-2012} {{SentencePiece: A simple and language independent subword tokenizer and detokenizer for Neural Text Processing}}.
\newblock In \emph{Proceedings of the 2018 Conference on Empirical Methods in Natural Language Processing: System Demonstrations}, pages 66--71, Brussels, Belgium. Association for Computational Linguistics.

\bibitem[{Li and Doddipatla(2023)}]{conformer-sota3}
Mohan Li and Rama Doddipatla. 2023.
\newblock \href {https://doi.org/10.1109/SLT54892.2023.10023042} {Non-autoregressive end-to-end approaches for joint automatic speech recognition and spoken language understanding}.
\newblock In \emph{2022 IEEE Spoken Language Technology Workshop (SLT)}, pages 390--397.

\bibitem[{Liskov(1975)}]{10.1145/987305.987309}
Barbara~H. Liskov. 1975.
\newblock \href {https://doi.org/10.1145/987305.987309} {Data types and program correctness}.
\newblock \emph{SIGPLAN Not.}, 10(7):16–17.

\bibitem[{Liu et~al.(2020)Liu, Zhu, Zhang, and Zong}]{liu2020bridging}
Yuchen Liu, Junnan Zhu, Jiajun Zhang, and Chengqing Zong. 2020.
\newblock \href {http://arxiv.org/abs/2010.14920} {{Bridging the Modality Gap for Speech-to-Text Translation}}.

\bibitem[{Lu et~al.(2019)Lu, Li, He, Sun, Dong, Qin, Wang, and Liu}]{lu-et-al-2016-macaron-net}
Yiping Lu, Zhuohan Li, Di~He, Zhiqing Sun, Bin Dong, Tao Qin, Liwei Wang, and Tie-Yan Liu. 2019.
\newblock \href {https://doi.org/10.48550/ARXIV.1906.02762} {Understanding and improving transformer from a multi-particle dynamic system point of view}.

\bibitem[{Ma et~al.(2021)Ma, Petridis, and Pantic}]{ma2021end}
Pingchuan Ma, Stavros Petridis, and Maja Pantic. 2021.
\newblock End-to-end audio-visual speech recognition with conformers.
\newblock In \emph{ICASSP 2021-2021 IEEE International Conference on Acoustics, Speech and Signal Processing (ICASSP)}, pages 7613--7617. IEEE.

\bibitem[{Marie et~al.(2021)Marie, Fujita, and Rubino}]{marie-etal-2021-scientific}
Benjamin Marie, Atsushi Fujita, and Raphael Rubino. 2021.
\newblock \href {https://doi.org/10.18653/v1/2021.acl-long.566} {Scientific credibility of machine translation research: A meta-evaluation of 769 papers}.
\newblock In \emph{Proceedings of the 59th Annual Meeting of the Association for Computational Linguistics and the 11th International Joint Conference on Natural Language Processing (Volume 1: Long Papers)}, pages 7297--7306, Online.

\bibitem[{McCall et~al.(1977)McCall, Richards, and Walters}]{McCall1977FactorsIS}
Jim~A. McCall, Paul~A. Richards, and Gene~F. Walters. 1977.
\newblock \emph{Factors in software quality}.
\newblock Rome Air Development Center, Rome.

\bibitem[{McCullough et~al.(2008)McCullough, McGeary, and Harrison}]{McCullough-2008-replicable}
Bruce~D. McCullough, Kerry~A. McGeary, and Teresa~D. Harrison. 2008.
\newblock \href {http://www.jstor.org/stable/25478330} {Do economics journal archives promote replicable research?}
\newblock \emph{The Canadian Journal of Economics}, 41(4):1406--1420.

\bibitem[{Narang et~al.(2021)Narang, Chung, Tay, Fedus, Fevry, Matena, Malkan, Fiedel, Shazeer, Lan, Zhou, Li, Ding, Marcus, Roberts, and Raffel}]{narang-etal-2021-transformer}
Sharan Narang, Hyung~Won Chung, Yi~Tay, Liam Fedus, Thibault Fevry, Michael Matena, Karishma Malkan, Noah Fiedel, Noam Shazeer, Zhenzhong Lan, Yanqi Zhou, Wei Li, Nan Ding, Jake Marcus, Adam Roberts, and Colin Raffel. 2021.
\newblock \href {https://doi.org/10.18653/v1/2021.emnlp-main.465} {Do transformer modifications transfer across implementations and applications?}
\newblock In \emph{Proceedings of the 2021 Conference on Empirical Methods in Natural Language Processing}, pages 5758--5773, Online and Punta Cana, Dominican Republic.

\bibitem[{Papi et~al.(2021)Papi, Gaido, Negri, and Turchi}]{papi-etal-2021-speechformer}
Sara Papi, Marco Gaido, Matteo Negri, and Marco Turchi. 2021.
\newblock \href {https://doi.org/10.18653/v1/2021.emnlp-main.127} {Speechformer: Reducing information loss in direct speech translation}.
\newblock In \emph{Proceedings of the 2021 Conference on Empirical Methods in Natural Language Processing}, pages 1698--1706, Online and Punta Cana, Dominican Republic. Association for Computational Linguistics.

\bibitem[{Park et~al.(2019)Park, Chan, Zhang, Chiu, Zoph, Cubuk, and Le}]{Park2019}
Daniel~S. Park, William Chan, Yu~Zhang, Chung-Cheng Chiu, Barret Zoph, Ekin~D. Cubuk, and Quoc~V. Le. 2019.
\newblock \href {https://doi.org/10.21437/Interspeech.2019-2680} {{SpecAugment: A Simple Data Augmentation Method for Automatic Speech Recognition}}.
\newblock In \emph{Proc. Interspeech 2019}, pages 2613--2617.

\bibitem[{Paszke et~al.(2019)Paszke, Gross, Massa, Lerer, Bradbury, Chanan, Killeen, Lin, Gimelshein, Antiga, Desmaison, K\"{o}pf, Yang, DeVito, Raison, Tejani, Chilamkurthy, Steiner, Fang, Bai, and Chintala}]{Paszke-2019-pytorch}
Adam Paszke, Sam Gross, Francisco Massa, Adam Lerer, James Bradbury, Gregory Chanan, Trevor Killeen, Zeming Lin, Natalia Gimelshein, Luca Antiga, Alban Desmaison, Andreas K\"{o}pf, Edward Yang, Zach DeVito, Martin Raison, Alykhan Tejani, Sasank Chilamkurthy, Benoit Steiner, Lu~Fang, Junjie Bai, and Soumith Chintala. 2019.
\newblock \emph{PyTorch: An Imperative Style, High-Performance Deep Learning Library}. Curran Associates Inc., Red Hook, NY, USA.

\bibitem[{Peng(2011)}]{Peng-2011-reproducible}
Roger~D. Peng. 2011.
\newblock \href {https://doi.org/10.1126/science.1213847} {Reproducible research in computational science}.
\newblock \emph{Science}, 334(6060):1226--1227.

\bibitem[{Perry and Wolf(1992)}]{10.1145/141874.141884}
Dewayne~E. Perry and Alexander~L. Wolf. 1992.
\newblock \href {https://doi.org/10.1145/141874.141884} {Foundations for the study of software architecture}.
\newblock \emph{SIGSOFT Softw. Eng. Notes}, 17(4):40–52.

\bibitem[{Pineau et~al.(2021)Pineau, Vincent-Lamarre, Sinha, Larivi{\`e}re, Beygelzimer, d’Alch{\'e} Buc, Fox, and Larochelle}]{pineau2021improving}
Joelle Pineau, Philippe Vincent-Lamarre, Koustuv Sinha, Vincent Larivi{\`e}re, Alina Beygelzimer, Florence d’Alch{\'e} Buc, Emily Fox, and Hugo Larochelle. 2021.
\newblock Improving reproducibility in machine learning research: a report from the neurips 2019 reproducibility program.
\newblock \emph{Journal of Machine Learning Research}, 22.

\bibitem[{Post(2018)}]{post-2018-call}
Matt Post. 2018.
\newblock \href {https://www.aclweb.org/anthology/W18-6319} {{A Call for Clarity in Reporting {BLEU} Scores}}.
\newblock In \emph{Proceedings of the Third Conference on Machine Translation: Research Papers}, pages 186--191, Belgium, Brussels.

\bibitem[{Prinz et~al.(2011)Prinz, Schlange, and Asadullah}]{prinz2011believe}
Florian Prinz, Thomas Schlange, and Khusru Asadullah. 2011.
\newblock Believe it or not: how much can we rely on published data on potential drug targets?
\newblock \emph{Nature reviews Drug discovery}, 10(9):712--712.

\bibitem[{Raff(2019)}]{NEURIPS2019_c429429b}
Edward Raff. 2019.
\newblock \href {https://proceedings.neurips.cc/paper/2019/file/c429429bf1f2af051f2021dc92a8ebea-Paper.pdf} {A step toward quantifying independently reproducible machine learning research}.
\newblock In \emph{Advances in Neural Information Processing Systems}, volume~32. Curran Associates, Inc.

\bibitem[{Ramachandran et~al.(2017)Ramachandran, Zoph, and Le}]{swish-2017}
Prajit Ramachandran, Barret Zoph, and Quoc~V. Le. 2017.
\newblock \href {https://doi.org/10.48550/ARXIV.1710.05941} {Searching for activation functions}.

\bibitem[{Ravanelli et~al.(2021)Ravanelli, Parcollet, Plantinga, Rouhe, Cornell, Lugosch, Subakan, Dawalatabad, Heba, Zhong, Chou, Yeh, Fu, Liao, Rastorgueva, Grondin, Aris, Na, Gao, Mori, and Bengio}]{speechbrain}
Mirco Ravanelli, Titouan Parcollet, Peter Plantinga, Aku Rouhe, Samuele Cornell, Loren Lugosch, Cem Subakan, Nauman Dawalatabad, Abdelwahab Heba, Jianyuan Zhong, Ju-Chieh Chou, Sung-Lin Yeh, Szu-Wei Fu, Chien-Feng Liao, Elena Rastorgueva, François Grondin, William Aris, Hwidong Na, Yan Gao, Renato~De Mori, and Yoshua Bengio. 2021.
\newblock \href {http://arxiv.org/abs/2106.04624} {{SpeechBrain}: A general-purpose speech toolkit}.
\newblock ArXiv:2106.04624.

\bibitem[{Ribeiro et~al.(2020)Ribeiro, Wu, Guestrin, and Singh}]{ribeiro-etal-2020-beyond}
Marco~Tulio Ribeiro, Tongshuang Wu, Carlos Guestrin, and Sameer Singh. 2020.
\newblock \href {https://doi.org/10.18653/v1/2020.acl-main.442} {Beyond accuracy: Behavioral testing of {NLP} models with {C}heck{L}ist}.
\newblock In \emph{Proceedings of the 58th Annual Meeting of the Association for Computational Linguistics}, pages 4902--4912, Online. Association for Computational Linguistics.

\bibitem[{Rogers et~al.(2021)Rogers, Baldwin, and Leins}]{rogers-etal-2021-just-think}
Anna Rogers, Timothy Baldwin, and Kobi Leins. 2021.
\newblock \href {https://doi.org/10.18653/v1/2021.findings-emnlp.414} {{{`}Just What do You Think You{'}re Doing, Dave?{'} A Checklist for Responsible Data Use in {NLP}}}.
\newblock In \emph{Findings of the Association for Computational Linguistics: EMNLP 2021}, pages 4821--4833, Punta Cana, Dominican Republic. Association for Computational Linguistics.

\bibitem[{Rozier and Rozier(2014)}]{Rozier-2014-repro}
Kristin~Y. Rozier and Eric W.~D. Rozier. 2014.
\newblock \href {https://doi.org/10.1109/ETHICS.2014.6893384} {Reproducibility, correctness, and buildability: The three principles for ethical public dissemination of computer science and engineering research}.
\newblock In \emph{2014 IEEE International Symposium on Ethics in Science, Technology and Engineering}, pages 1--13.

\bibitem[{Sadowski et~al.(2018)Sadowski, S\"{o}derberg, Church, Sipko, and Bacchelli}]{10.1145/3183519.3183525}
Caitlin Sadowski, Emma S\"{o}derberg, Luke Church, Michal Sipko, and Alberto Bacchelli. 2018.
\newblock \href {https://doi.org/10.1145/3183519.3183525} {Modern code review: A case study at google}.
\newblock In \emph{Proceedings of the 40th International Conference on Software Engineering: Software Engineering in Practice}, ICSE-SEIP '18, page 181–190, New York, NY, USA. Association for Computing Machinery.

\bibitem[{Schloss(2018)}]{doi:10.1128/mBio.00525-18}
Patrick~D. Schloss. 2018.
\newblock \href {https://doi.org/10.1128/mBio.00525-18} {Identifying and overcoming threats to reproducibility, replicability, robustness, and generalizability in microbiome research}.
\newblock \emph{mBio}, 9(3):e00525--18.

\bibitem[{Schwartz et~al.(2020)Schwartz, Dodge, Smith, and Etzioni}]{greenai}
Roy Schwartz, Jesse Dodge, Noah~A. Smith, and Oren Etzioni. 2020.
\newblock \href {https://doi.org/10.1145/3381831} {Green ai}.
\newblock \emph{Commun. ACM}, 63(12):54–63.

\bibitem[{Shterionov and Vanmassenhove(2023)}]{Shterionov2023}
Dimitar Shterionov and Eva Vanmassenhove. 2023.
\newblock \href {https://doi.org/10.1007/978-3-031-14689-3_10} {{The Ecological Footprint of Neural Machine Translation Systems}}.
\newblock In Helena Moniz and Carla Parra~Escart{\'i}n, editors, \emph{Towards Responsible Machine Translation: Ethical and Legal Considerations in Machine Translation}, pages 185--213. Springer International Publishing, Cham, Switzerland.

\bibitem[{Srivastava et~al.(2014)Srivastava, Hinton, Krizhevsky, Sutskever, and Salakhutdinov}]{Srivastava-2014-dropout}
Nitish Srivastava, Geoffrey Hinton, Alex Krizhevsky, Ilya Sutskever, and Ruslan Salakhutdinov. 2014.
\newblock Dropout: A simple way to prevent neural networks from overfitting.
\newblock \emph{J. Mach. Learn. Res.}, 15(1):1929–1958.

\bibitem[{Srivastava et~al.(2022)Srivastava, Wang, Tjandra, Kumar, Liu, Singh, and Saraf}]{conformer-sota2}
Sangeeta Srivastava, Yun Wang, Andros Tjandra, Anurag Kumar, Chunxi Liu, Kritika Singh, and Yatharth Saraf. 2022.
\newblock \href {https://doi.org/10.1109/ICASSP43922.2022.9746490} {Conformer-based self-supervised learning for non-speech audio tasks}.
\newblock In \emph{ICASSP 2022 - 2022 IEEE International Conference on Acoustics, Speech and Signal Processing (ICASSP)}, pages 8862--8866.

\bibitem[{Strubell et~al.(2019)Strubell, Ganesh, and McCallum}]{strubell-etal-2019-energy}
Emma Strubell, Ananya Ganesh, and Andrew McCallum. 2019.
\newblock \href {https://doi.org/10.18653/v1/P19-1355} {Energy and policy considerations for deep learning in {NLP}}.
\newblock In \emph{Proceedings of the 57th Annual Meeting of the Association for Computational Linguistics}, pages 3645--3650, Florence, Italy. Association for Computational Linguistics.

\bibitem[{{The Turing Way Community}(2022)}]{the_turing_way_community_2022}
{The Turing Way Community}. 2022.
\newblock \href {https://doi.org/0.5281/zenodo.3233853} {{The Turing Way: A Handbook for Reproducible Data Science}}.

\bibitem[{Tripathy and Naik(2011)}]{tripathy2011software}
Priyadarshi Tripathy and Kshirasagar Naik. 2011.
\newblock \emph{Software testing and quality assurance: theory and practice}.
\newblock John Wiley \& Sons.

\bibitem[{Trisovic et~al.(2022)Trisovic, Lau, Pasquier, and Crosas}]{Trisovic2022}
Ana Trisovic, Matthew~K. Lau, Thomas Pasquier, and Merc{\`e} Crosas. 2022.
\newblock \href {https://doi.org/10.1038/s41597-022-01143-6} {A large-scale study on research code quality and execution}.
\newblock \emph{Scientific Data}, 9(1):60.

\bibitem[{Tsiamas et~al.(2023)Tsiamas, Gállego, Fonollosa, and Costa-jussà}]{Tsiamas-2023-Perceivers}
Ioannis Tsiamas, Gerard~I. Gállego, José A.~R. Fonollosa, and Marta~R. Costa-jussà. 2023.
\newblock \href {https://doi.org/10.1109/ICASSP49357.2023.10095276} {{Efficient Speech Translation with Dynamic Latent Perceivers}}.
\newblock In \emph{ICASSP 2023 - 2023 IEEE International Conference on Acoustics, Speech and Signal Processing (ICASSP)}, pages 1--5.

\bibitem[{Ulmer et~al.(2022)Ulmer, Bassignana, M{\"u}ller-Eberstein, Varab, Zhang, Hardmeier, and Plank}]{ulmer2022experimental}
Dennis Ulmer, Elisa Bassignana, Max M{\"u}ller-Eberstein, Daniel Varab, Mike Zhang, Christian Hardmeier, and Barbara Plank. 2022.
\newblock Experimental standards for deep learning research: A natural language processing perspective.
\newblock \emph{arXiv preprint arXiv:2204.06251}.

\bibitem[{Vaswani et~al.(2017)Vaswani, Shazeer, Parmar, Uszkoreit, Jones, Gomez, Kaiser, and Polosukhin}]{NIPS2017_3f5ee243}
Ashish Vaswani, Noam Shazeer, Niki Parmar, Jakob Uszkoreit, Llion Jones, Aidan~N Gomez, \L~ukasz Kaiser, and Illia Polosukhin. 2017.
\newblock \href {https://proceedings.neurips.cc/paper/2017/file/3f5ee243547dee91fbd053c1c4a845aa-Paper.pdf} {Attention is all you need}.
\newblock In \emph{Advances in Neural Information Processing Systems}, volume~30. Curran Associates, Inc.

\bibitem[{Wang et~al.(2020)Wang, Tang, Ma, Wu, Okhonko, and Pino}]{wang2020fairseqs2t}
Changhan Wang, Yun Tang, Xutai Ma, Anne Wu, Dmytro Okhonko, and Juan Pino. 2020.
\newblock fairseq s2t: Fast speech-to-text modeling with fairseq.
\newblock In \emph{Proceedings of the 2020 Conference of the Asian Chapter of the Association for Computational Linguistics (AACL): System Demonstrations}.

\bibitem[{Weiss et~al.(2017)Weiss, Chorowski, Jaitly, Wu, and Chen}]{weiss2017sequence}
Ron~J. Weiss, Jan Chorowski, Navdeep Jaitly, Yonghui Wu, and Zhifeng Chen. 2017.
\newblock {Sequence-to-Sequence Models Can Directly Translate Foreign Speech}.
\newblock In \emph{Proceedings of Interspeech 2017}, pages 2625--2629, Stockholm, Sweden.

\bibitem[{Wieling et~al.(2018{\natexlab{a}})Wieling, Rawee, and van Noord}]{10.1162/coli_a_00330}
Martijn Wieling, Josine Rawee, and Gertjan van Noord. 2018{\natexlab{a}}.
\newblock \href {https://doi.org/10.1162/coli_a_00330} {{Reproducibility in Computational Linguistics: Are We Willing to Share?}}
\newblock \emph{Computational Linguistics}, 44(4):641--649.

\bibitem[{Wieling et~al.(2018{\natexlab{b}})Wieling, Rawee, and van Noord}]{wieling-etal-2018-squib}
Martijn Wieling, Josine Rawee, and Gertjan van Noord. 2018{\natexlab{b}}.
\newblock \href {https://doi.org/10.1162/coli_a_00330} {{S}quib: Reproducibility in computational linguistics: Are we willing to share?}
\newblock \emph{Computational Linguistics}, 44(4):641--649.

\bibitem[{Williams et~al.(2003)Williams, Maximilien, and Vouk}]{Williams-2003-TDD}
Laurie Williams, E.~Michael Maximilien, and Mladen Vouk. 2003.
\newblock \href {https://doi.org/10.1109/ISSRE.2003.1251029} {Test-driven development as a defect-reduction practice}.
\newblock In \emph{14th International Symposium on Software Reliability Engineering, 2003. ISSRE 2003.}, pages 34--45.

\bibitem[{Xu et~al.(2021)Xu, Hu, Li, Zhang, Huang, Ju, Xiao, and Zhu}]{xu-etal-2021-stacked}
Chen Xu, Bojie Hu, Yanyang Li, Yuhao Zhang, Shen Huang, Qi~Ju, Tong Xiao, and Jingbo Zhu. 2021.
\newblock \href {https://doi.org/10.18653/v1/2021.acl-long.204} {Stacked acoustic-and-textual encoding: Integrating the pre-trained models into speech translation encoders}.
\newblock In \emph{Proceedings of the 59th Annual Meeting of the Association for Computational Linguistics and the 11th International Joint Conference on Natural Language Processing (Volume 1: Long Papers)}, pages 2619--2630, Online.

\bibitem[{Yang et~al.(2021)Yang, Hira, Ni, Chourdia, Astafurov, Chen, Yeh, Puhrsch, Pollack, Genzel, Greenberg, Yang, Lian, Mahadeokar, Hwang, Chen, Goldsborough, Roy, Narenthiran, Watanabe, Chintala, Quenneville-Bélair, and Shi}]{yang2021torchaudio}
Yao-Yuan Yang, Moto Hira, Zhaoheng Ni, Anjali Chourdia, Artyom Astafurov, Caroline Chen, Ching-Feng Yeh, Christian Puhrsch, David Pollack, Dmitriy Genzel, Donny Greenberg, Edward~Z. Yang, Jason Lian, Jay Mahadeokar, Jeff Hwang, Ji~Chen, Peter Goldsborough, Prabhat Roy, Sean Narenthiran, Shinji Watanabe, Soumith Chintala, Vincent Quenneville-Bélair, and Yangyang Shi. 2021.
\newblock Torchaudio: Building blocks for audio and speech processing.
\newblock \emph{arXiv preprint arXiv:2110.15018}.

\bibitem[{Zhang et~al.(2022)Zhang, Haddow, and Sennrich}]{pmlr-v162-zhang22i}
Biao Zhang, Barry Haddow, and Rico Sennrich. 2022.
\newblock \href {https://proceedings.mlr.press/v162/zhang22i.html} {Revisiting end-to-end speech-to-text translation from scratch}.
\newblock In \emph{Proceedings of the 39th International Conference on Machine Learning}, volume 162 of \emph{Proceedings of Machine Learning Research}, pages 26193--26205. PMLR.

\bibitem[{Zhao et~al.(2021)Zhao, Luo, Chen, and Gilman}]{zhao-etal-2021-mutual}
Jiawei Zhao, Wei Luo, Boxing Chen, and Andrew Gilman. 2021.
\newblock \href {https://doi.org/10.18653/v1/2021.emnlp-main.325} {Mutual-learning improves end-to-end speech translation}.
\newblock In \emph{Proceedings of the 2021 Conference on Empirical Methods in Natural Language Processing}, pages 3989--3994, Online and Punta Cana, Dominican Republic.

\end{thebibliography}
